\documentclass[10pt,twocolumn,letterpaper]{article}

\usepackage[pagenumbers]{cvpr} 

\usepackage{times}
\usepackage{epsfig}
\usepackage{multirow}
\usepackage{multicol}
\usepackage{caption}
\usepackage{mathtools}
\usepackage{eufrak}
\usepackage{url}
\usepackage{lipsum}
\usepackage{graphicx}
\usepackage{amsmath}
\usepackage{amssymb}
\usepackage{booktabs}
\usepackage{enumitem}
\usepackage{filecontents}
\usepackage[italian,english]{babel}
\usepackage[autostyle]{csquotes}  
\usepackage[dvipsnames]{xcolor}

\usepackage{xcolor}
\definecolor{tablegray}{gray}{0.6}

\usepackage[pagebackref,breaklinks,colorlinks]{hyperref}

\usepackage[capitalize]{cleveref}
\crefname{section}{Sec.}{Secs.}
\Crefname{section}{Section}{Sections}
\Crefname{table}{Table}{Tables}
\crefname{table}{Tab.}{Tabs.}

\def\cvprPaperID{5303} 
\def\confName{CVPR}
\def\confYear{2022}

\begin{document}

\title{Advancing High-Resolution Video-Language Representation \\ with Large-Scale Video Transcriptions}


\author{Hongwei Xue\thanks{Equal contribution in alphabetical order. This work was performed when Hongwei Xue, Tiankai Hang, Yanhong Zeng and Yuchong Sun were visiting Microsoft Research Asia as research interns. Corresponding authors: Bei Liu, Huan Yang, Jianlong Fu.},
Tiankai Hang\footnotemark[1],
Yanhong Zeng\footnotemark[1],
Yuchong Sun\footnotemark[1],\\
Bei Liu,
Huan Yang,
Jianlong Fu,
Baining Guo\\
Microsoft Research Asia \\
{\tt\small \{v-honxue,v-tiahang,t-yazen,v-yuchongsun,bei.liu,huayan,jianf,bainguo\}@microsoft.com}
}

\maketitle

\newcommand{\figman}{
\begin{figure}[t]
 \centering
 \includegraphics[width=\linewidth,trim=15 0 10 0]{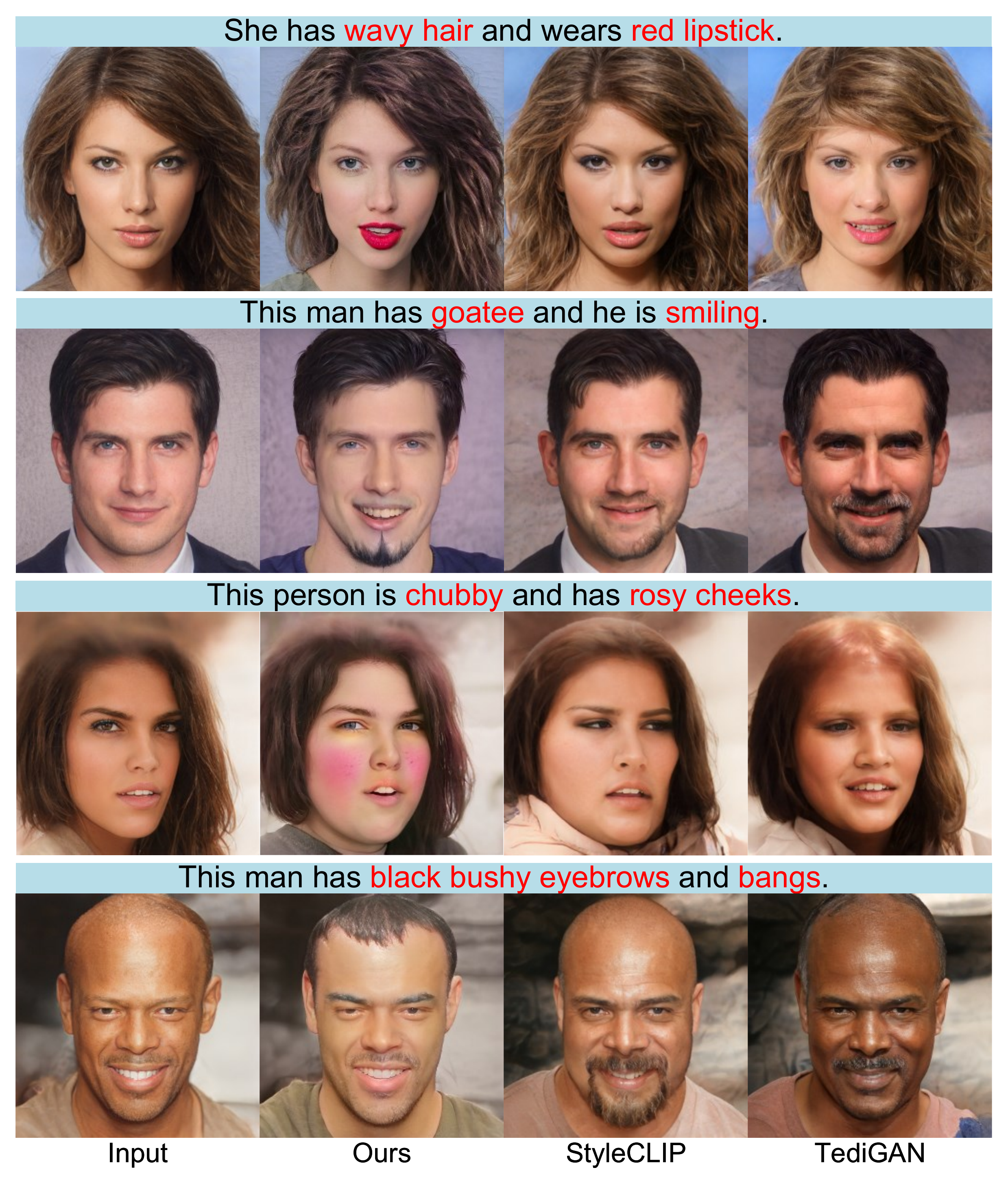}
 \vspace{-8mm}
 \caption{Text-guided manipulation compared with StyleCLIP \cite{patashnik2021styleclip} and TediGAN \cite{xia2021tedigan}.  Our model is able to handle complex descriptions and edit the inputs according to the target attributes (highlighted in {\color{red}red}) better. All the inputs are of \(1024\times 1024\) size.}
 \label{fig:man}
 \vspace{-5mm}
\end{figure}
}

\newcommand{\figsr}{
\begin{figure}[t]
 \centering
 \includegraphics[width=\linewidth,trim=15 0 10 0]{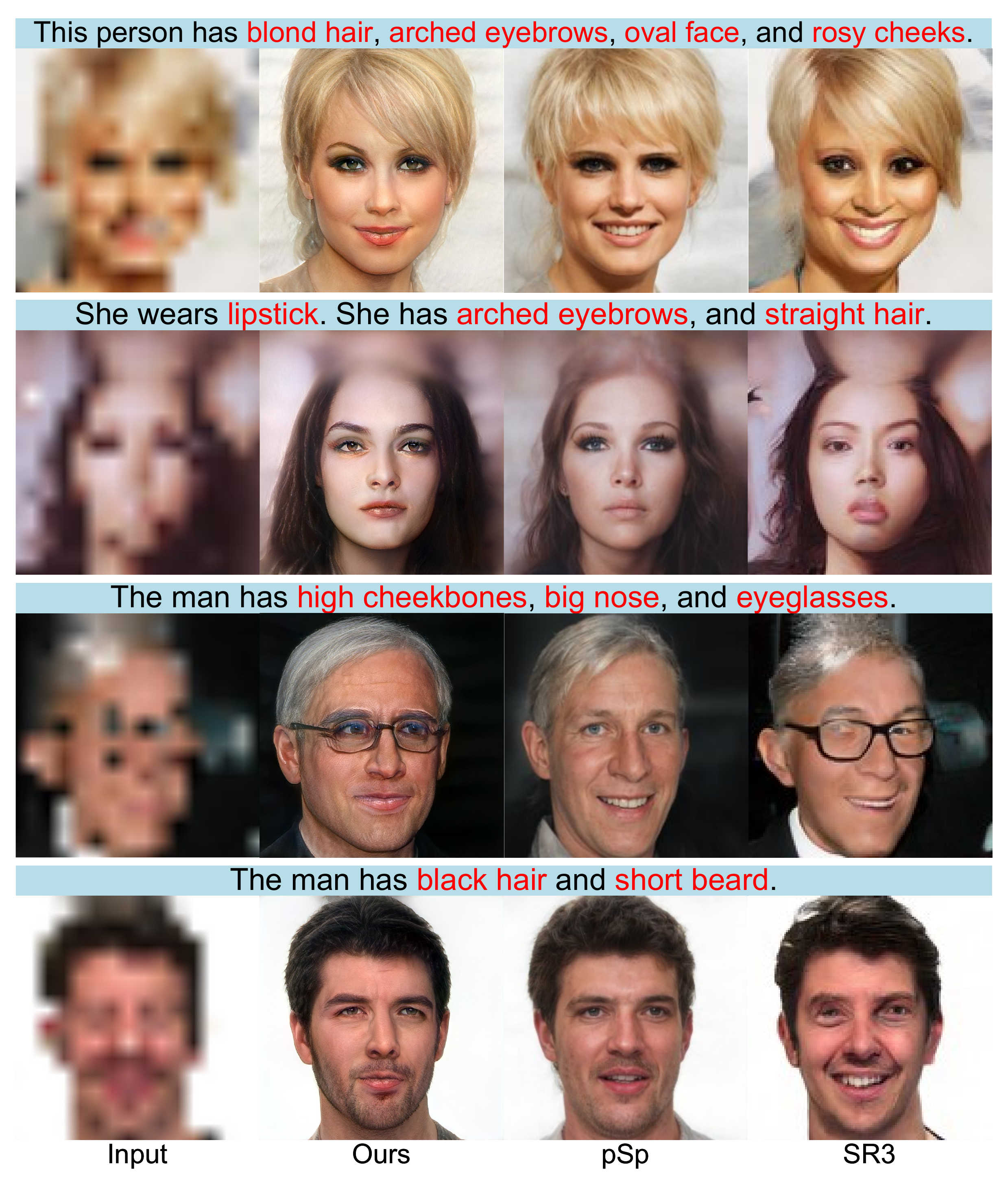}
 \vspace{-8mm}
 \caption{Text-guided super-resolution compared with pSp \cite{richardson2021psp} and SR3 \cite{saharia2021sr3}. Our model is able to reconstruct more accurate target attributes with descriptions (\textit{e.g.}, eyeglasses in the third case). All inputs are upsampled from \(16\times 16\) to \(1024\times 1024\).}
 \label{fig:sr}
 \vspace{-5mm}
\end{figure}
}

\begin{abstract}

We study joint video and language (VL) pre-training to enable cross-modality learning and benefit plentiful downstream VL tasks. Existing works either extract low-quality video features or learn limited text embedding, while neglecting that high-resolution videos and diversified semantics can significantly improve cross-modality learning. In this paper, we propose a novel \textbf{H}igh-resolution and \textbf{D}iversified \textbf{VI}deo-\textbf{LA}nguage pre-training model (HD-VILA) for many visual tasks. In particular, we collect a large dataset with two distinct properties: 1) the first high-resolution dataset including 371.5k hours of 720p videos, and 2) the most diversified dataset covering 15 popular YouTube categories. To enable VL pre-training, we jointly optimize the HD-VILA model by a hybrid Transformer that learns rich spatiotemporal features, and a multimodal Transformer that enforces interactions of the learned video features with diversified texts. Our pre-training model achieves new state-of-the-art results in \textbf{10} VL understanding tasks and \textbf{2} more novel text-to-visual generation tasks. For example, we outperform SOTA models with relative increases of 40.4\% R@1 in zero-shot MSR-VTT text-to-video retrieval task, and 55.4\% in high-resolution dataset LSMDC. The learned VL embedding is also effective in generating visually pleasing and semantically relevant results in text-to-visual editing and super-resolution tasks. 

\end{abstract}
\vspace{-1mm}
\section{Introduction} \label{introduction}

Recent years we have witnessed an increasing number of videos with the popularity of appealing video websites and mobile apps (\textit{e.g.}, YouTube, TikTok). 
As the rapid development of smartphone cameras, device storage, and 5G networks, high-quality video creation, and diverse content sharing like travel, sports, and music become a new fashion. Therefore, the capability of video analytic and joint high-level understanding with language play a key role in many video tasks, such as video search~\cite{bain2021frozen, patrick2021supportset}, video recommendation~\cite{TaoVideoRec_MM12}, and video editing \cite{xia2021tedigan, patashnik2021styleclip}. To facilitate video understanding, we study joint video and language (VL) pre-training, which is a new paradigm in both natural language processing \cite{Devlin2018} and computer vision \cite{xu2021videoclip,huang2021seeing}.

Existing video-language understanding models are highly limited in either scale or scope of video-language datasets. Early datasets (\textit{e.g.}, MSR-VTT \cite{xu2016msr}, DiDeMo \cite{anne2017localizing}) consist of videos and descriptions that are manually annotated by humans. The heavy and expensive annotation cost limits the scale of data. Moreover, datasets with only descriptive sentences are limited in complexity and variability that largely hinders generalization power. Recently, several datasets \cite{bain2021frozen,miech2019howto100m} are proposed by transcriptions along with videos using ASR (automatic speech recognition), so that the data scale can be greatly enlarged. One most representative work is HowTo100M \cite{miech2019howto100m} which consists of million-scale instructional videos. However, there are still large gaps between these video datasets and real-scenario videos in terms of video quality and semantic diversity.

\begin{figure*}
    \centering
    \includegraphics[width=0.95\textwidth]{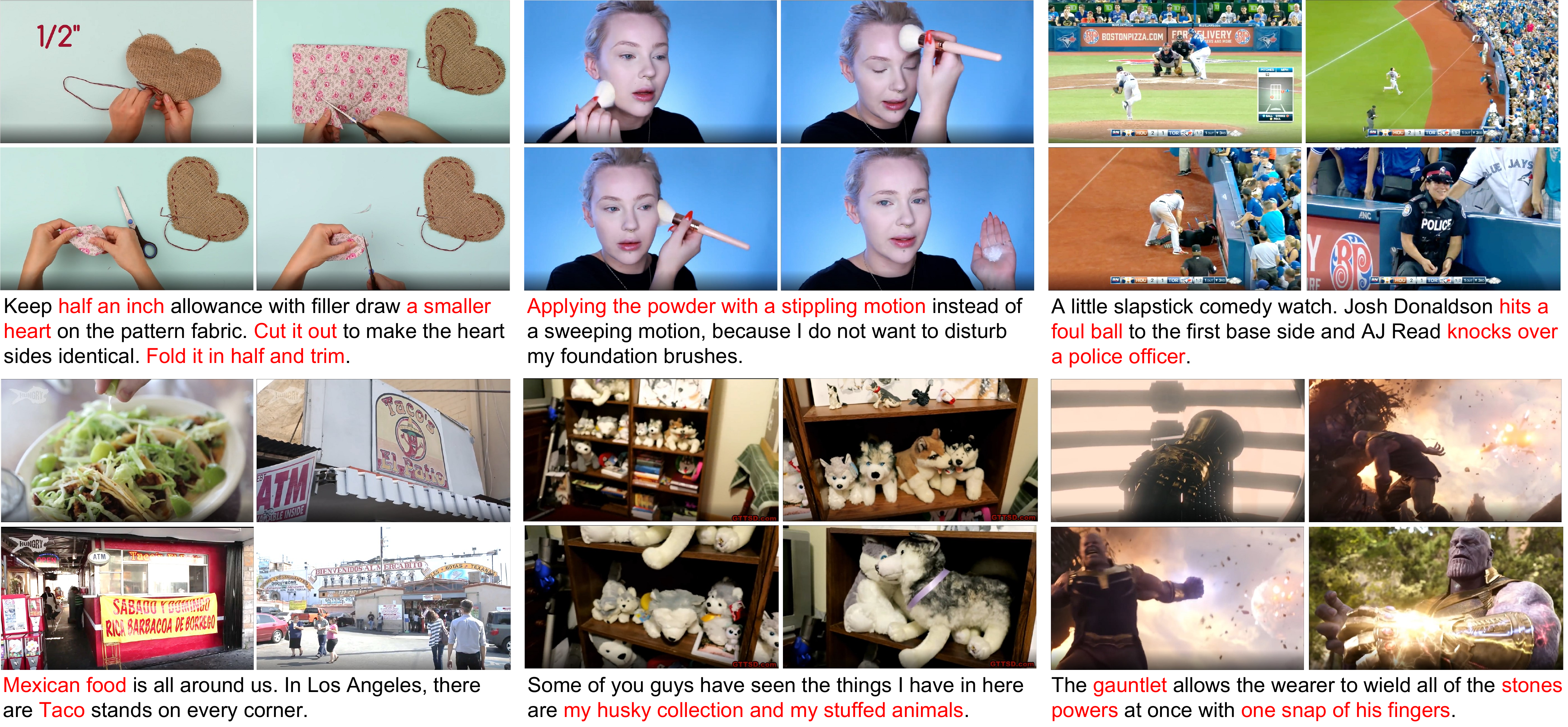}
    \vspace{-3mm}
    \caption{Examples of video clips and ASR generated transcriptions in the proposed HD-VILA-100M dataset. We present six samples (four frames for each), with diverse video categories covering HowTo \& Style, People \& Blog, Sports, Travel \& Event, Pets \& Animals, Film \& Animation. Relevant words from auto-generated video transcriptions are manually highlighted in red. [Best viewed in Color]}
    \label{fig:dataset_example}
    \vspace{-5mm}
\end{figure*}

To tackle the above limitations, we propose the HD-VILA-100M dataset (\textit{i.e.}, \textbf{H}igh-resolution and \textbf{D}iversified \textbf{VI}deo and \textbf{LA}nguage) which covers a wide range of video categories and benefits a plenty of VL tasks, such as text-to-video retrieval \cite{patrick2021supportset} and video QA \cite{lei2021less}. This dataset has the following key properties: (1) Large: we have collected one of the largest video-language datasets, which consists of 100M video clip and sentence pairs from 3.3 million videos with 371.5K hours in total ($2.8\times$ video hour and $8\times$ average sentence length than HowTo100M~\cite{miech2019howto100m}). 
(2) High resolution: all the videos are 720p which is much higher quality than existing datasets that are mostly 240p or 360p.
(3) Diverse and balanced: we cover a wide range of topics from the YouTube, with 15 popular categories (\textit{e.g.}, sports, music, autos). Meanwhile, we ensure a balanced video clip number in each category to ease underfit problem.

To enable video-language pre-training, effective video representation is essential. 
Due to computational limitations (\textit{e.g.}, memory), previous works either 1) adopt simple frame-based encoders and turn to end-to-end visual encoding and multimodal fusion \cite{lei2021less}, or 2) choose advanced spatiotemporal encoders \cite{carreira2017quo,xie2018rethinking}, while having to do visual encoding and multimodal fusion step-by-step. Few works can learn joint spatiotemporal video representation with end-to-end video-language pre-training. 

In this paper, we propose to utilize \textbf{hybrid image sequence} that consists of few high-resolution (HR) frames and more low-resolution (LR) neighbor frames for multiple video learning tasks. Such a design enables end-to-end training with high-resolution spatiotemporal video representation. To achieve this goal, we tackle two questions:  
(1) Which HR and LR frames should be sampled?
(2) How to learn spatiotemporal features with the hybrid image sequences?
For the first problem, we randomly sample HR frames from a video clip to ensure the robustness of learned video features. LR frames are uniformly sampled from its surroundings considering that neighboring frames contain similar spatial information and are critical to temporal feature learning.
Second, we propose to encode HR and LR frames separately while mapping HR feature to a joint embedding space with LR features by a hybrid Transformer. Such design ensures the spatiotemporal representation of videos to cover both HR and LR frames in a learnable way.
The learned spatiotemporal feature is further combined with detailed spatial features, followed by a multimodal Transformer that learns to optimize video and language embedding in an end-to-end manner.

Our contributions are summarized as follows: 1) We use automatic video transcriptions to build to-date the largest high-resolution and diversified video-language dataset; 2) We propose a novel pre-training framework to learn spatiotemporal information for video representation from hybrid image sequences that consist of HR and LR frames; 3) Extensive experiments verify the effectiveness of the learned cross-modality embedding in \textbf{10} video understanding and \textbf{2} text-to-visual generation tasks.
The dataset, model and code are released \footnote{\url{https://github.com/microsoft/XPretrain/tree/main/hd-vila}}.
\begin{table*}[h]
\small
    \centering
    \begin{tabular}{l l r r r r r r} 
    \hline
    Dataset & Domain  & \#Video clips & \#Sentence & Avg len(sec) & Sent len & Duration(h) & Resolution \\\hline\hline
    MSR-VTT~\cite{xu2016msr} & open & 10K & 200K & 15.0 & 9.3 & 40 & 240p\\ 
    DideMo~\cite{anne2017localizing} & Flickr & 27K & 41K & 6.9 & 8.0 &87 & -\\ 
    
    LSMDC~\cite{Rohrbach2016MovieD} & movie & 118K & 118K & 4.8 & 7.0 & 158 & 1080p\\
    YouCook \uppercase\expandafter{\romannumeral2}~\cite{Zhou2018TowardsAL} & cooking & 14K & 14K & 19.6 & 8.8 & 176 & -\\
    How2~\cite{sanabria2018how2} & instructional & 80K & 80K & 90.0 & 20.0 & 2K & -\\
    ActivityNet Caption~\cite{Krishna2017actnetcaption} & action & 100K & 100K & 36.0 & 13.5 & 849 & -\\
    WebVid-2M~\cite{bain2021frozen} & open & 2.5M & 2.5M & 18.0 & 12.0 & 13K & 360p\\
    HowTo100M~\cite{miech2019howto100m} & instructional & 136M & 136M & 3.6 & 4.0 & 134.5K & 240p \\ \hline
    HD-VILA-100M (Ours) & open & 103M & 103M & 13.4 & 32.5 & 371.5K & 720p\\
    \hline
    \end{tabular}
    \vspace{-3mm}
    \caption{Statistics of HD-VILA-100M and its comparison with existing video-language datasets.}
    \label{tab:datasets}
    \vspace{-5mm}
\end{table*}

\vspace{-2mm}
\section{Related Work}
\vspace{-1mm}
\paragraph{Video Representation}
Video representation are typically designed with 2D/3D CNNs \cite{tran2015learning,carreira2017quo,xie2018rethinking} or Transformers \cite{gberta_2021_ICML}. 
Pioneering works of VL pre-training \cite{sun2019videobert,patrick2021supportset,zhu2020actbert} adopt pre-extracted video features (\textit{e.g.}, S3D \cite{zhang2018s3d}, I3D~\cite{carreira2017quo}) for video representation.
While in image-language pre-training, researchers find that end-to-end training will decrease the domain gap of visual representation and improve the generalization for image-text tasks \cite{huang2021seeing}. While for video representation, it is too heavy to make the video-based encoder (\textit{e.g.}, S3D, I3D, ResNet~\cite{he2016deep}, SlowFast~\cite{feichtenhofer2019slowfast}) trainable. Thus, some works \cite{lei2021less,zellers2021merlot} utilize the image-based encoder (\textit{e.g.}, ResNet~\cite{he2016deep}, ViT~\cite{dosovitskiy2020vit}) with a sparse sampling mechanism to make the visual encoder trainable. 
In this paper, we explore how to make a video encoder trainable in consideration of both spatial and temporal features.

\vspace{-3mm}
\paragraph{Video-Language Pre-Training}

Vision and language pre-training has attracted extensive attention in very recent years. Aligned with the success of image-language pre-training \cite{huang2020pixel,huang2021seeing,xue2021probing} and applications \cite{fu2014image,fu2015tagging,huang2021unifying,fu2017advances,li2019emotion,chen2019neural}, video-language pre-training is showing more and more promising potentials \cite{sun2019videobert,li2020hero,zhu2020actbert,tang2021decembert,lei2021less,patrick2021supportset}. Among them, some works concentrate on specific type of downstream tasks such as video-text retrieval \cite{bain2021frozen,xu2021videoclip} and video question answering \cite{zellers2021merlot}. In this paper, we explore to pre-train a generalized model on diversified and large-scale data to adapt to different video-language tasks.
Video-language pre-training tasks can be mainly categorized into two types: reconstructive, contrastive. 
Reconstructive methods \cite {sun2019videobert,zhu2020actbert,li2020hero,tang2021decembert} usually adopt an early fusion architecture and aim to reconstruct a masked part in the visual or textual domain. Typical pre-training tasks are masked language modeling (MLM), masked frame modeling (MFM), frame order modeling (FOM). Contrastive methods \cite{miech2020end,zellers2021merlot} are inspired by contrastive learning and target to learn video-text matching. 
In this paper, we combine these two types of objectives for the final target.

\vspace{-2mm}
\section{Dataset}
\vspace{-1mm}

To facilitate the multimodal representation learning, we collect HD-VILA-100M, a large-scale, high-resolution, and diversified video-language dataset. 

\vspace{-1mm}
\subsection{Video Collection}
\begin{figure}
    \centering
    \includegraphics[width=0.95\columnwidth]{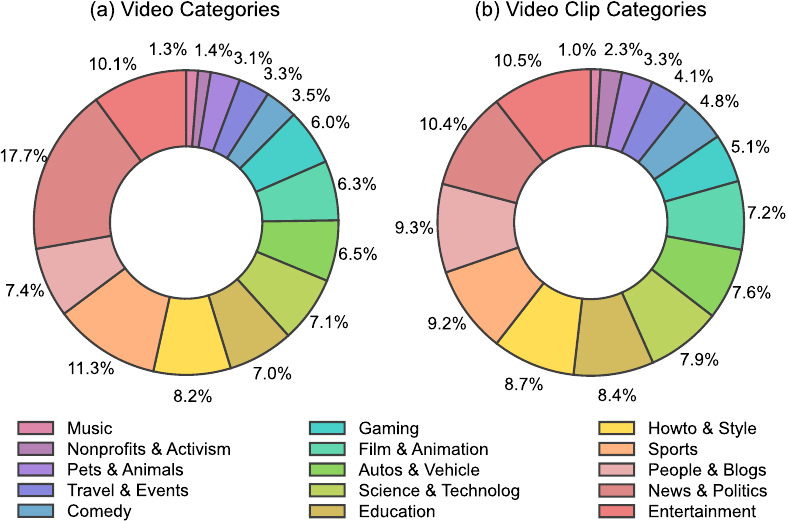}
    \vspace{-3mm}
    \caption{The distribution of categories in HD-VILA-100M dataset: (a) video, (b) video clip. [Best viewed in Color]}
    \label{fig:dataset_balance}
    \vspace{-6mm}
\end{figure}

\vspace{-1mm}
We choose YouTube as the video resource since it covers diverse categories of videos uploaded by different users, ranging from documentary films by professional TV channels to everyday vlogs by ordinary users. 
To cover more topics, we start from several official topics of YouTube videos.
To ensure the high quality of videos as well as better alignment of video and transcription, we search on the YouTube website and a video analysis website \footnote{\url{https://socialblade.com/youtube/}} to find popular YouTube channels, such as BBC Earth, National Geography, etc.
Videos in these channels and videos appeared in YouTube-8M \cite{abu2016youtube} and YT-Temporal-180M \cite{zellers2021merlot} make up a list of 14 million videos. 
We only keep videos with subtitles and 720p resolution. We then limit the time length of each category to 30K hours to avoid long tail. We only download videos with English transcripts. Finally, we obtain 3.3 million videos in total with high-quality and distributed in 15 categories in balance (as in Figure \ref{fig:dataset_balance}). 

\vspace{-1mm}
\subsection{Video Clip Selection and Text Processing}
\vspace{-1mm}
To effectively generate video-text pairs, we use transcriptions along with the videos as the language in HD-VILA-100M. Different from traditional video-language datasets \cite{xu2016msr,anne2017localizing} that use manual annotated descriptions for videos, transcriptions are available in large quantities and involve richer information.
However, many subtitles in YouTube videos are generated by ASR and are usually fragmentary and lacking punctuation. To split the subtitles for complete sentences, we utilize an off-the-shelf tool \footnote{\url{https://github.com/ottokart/punctuator2}} which shows 75.7\% accuracy on its test set. Then we make video clips by aligning the sentences to corresponding clips via Dynamic Time Warping using the timestamp of the original subtitles. After processing, each pair in the HD-VILA-100M consists of a video clip about 13.4 seconds on average and a sentence with 32.5 words on average.

\vspace{-1mm}
\subsection{Data Statistics}
\vspace{-1mm}
The detailed data statistics of HD-VILA-100M are listed in Table \ref{tab:datasets}. Compared with other video-language datasets, HD-VILA-100M is the largest video-language dataset in terms of duration and word number. More videos indicate richer visual information contained in HD-VILA-100M and longer sentences mean that the language includes more detailed and richer semantics. Compared with HowTo100M \cite{miech2019howto100m} which only includes instructional videos, HD-VILA-100M is derived from a wide range of domains and videos of each category is relatively balanced as shown in Figure \ref{fig:dataset_balance}. This merit can improve the generalization power of the pre-trained model. Moreover, all the videos in HD-VILA-100M are in 720p and the high quality ensures detailed information for video representation learning. In summary, HD-VILA-100M represents the largest, high-resolution, and diversified dataset for video and language learning.
\vspace{-2mm}
\section{Approach}
\vspace{-1mm}

\begin{figure}
    \centering
    \includegraphics[width=\columnwidth]{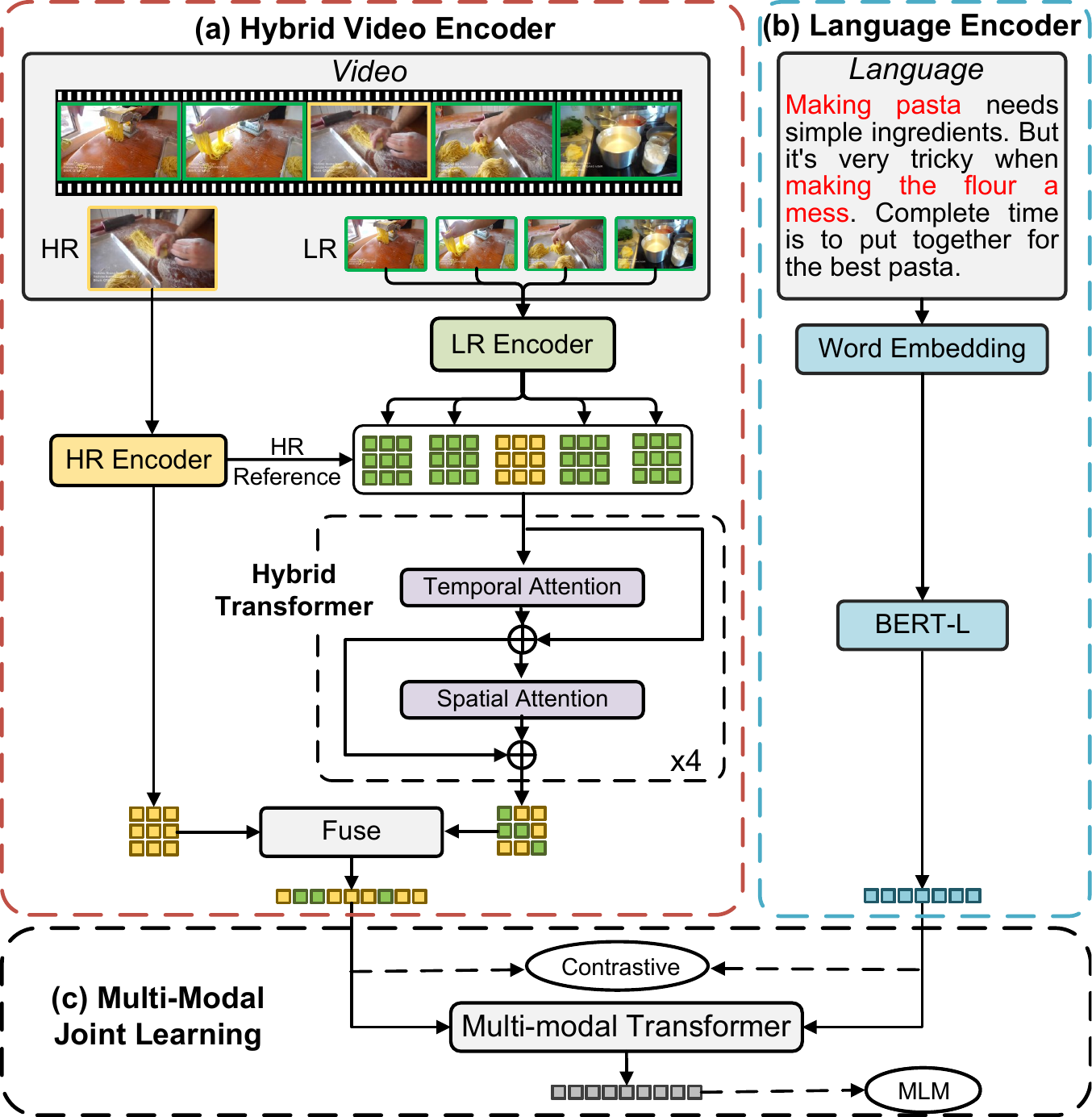}
    \vspace{-7mm}
    \caption{The framework of HD-VILA. Yellow and green colors indicate HR- and LR-related input, operation and output, respectively. Hybrid Transformer learns spatiotemporal representation from HR and LR features. [Best viewed in Color]}
    \label{fig:framework}
    \vspace{-6mm}
\end{figure}

Figure~\ref{fig:framework} shows the overall framework of \textbf{H}igh-resolution and \textbf{D}iversified \textbf{VI}deo-\textbf{LA}nguage (HD-VILA) model that consists of three parts: (a) hybrid video encoder, (b) language encoder, and (c) multi-modal joint learning.

\vspace{-1mm}
\subsection{Hybrid Video Encoder}\label{subsec:video_encoder}
\vspace{-1mm}
Since the video clips in our dataset are long-range with 13.4 seconds on average, we adopt the strategy of sparsely sampling a sequence of segments from a video clip and then aggregating their predictions similar to ClipBERT \cite{lei2021less}.
As explained in Section \ref{introduction}, for each segment $s_i$, we randomly takes one HR frame at $t$-th timestep $\boldsymbol{X}^{s_i}_{t} \in \mathbb{R}^{3 \times H \times W}$ and $2N$ surrounding LR frames 
$ \{\boldsymbol{X}^{s_i}_{t+kr} \in \mathbb{R}^{3 \times \frac{H}{4} \times \frac{W}{4}},k\in(-N,\ldots,-1,1,\ldots,N)\}$
to build a \textbf{hybrid image sequence}, where $r$ is LR frame sampling rate. 

In Figure~\ref{fig:framework}, the hybrid video encoder includes three parts: an HR encoder for HR frame, an LR encoder for LR neighbor frames and a \textbf{Hybrid Transformer} $T_{\mathrm{hy}}$ that learns spatiotemporal features by self-attention. 
HR encoder consists of a 4-stage ResNet $F_{\mathrm{hr}}$ and an adapter $D_{\mathrm{hr}}$. LR encoder is a 3-stage ResNet $F_{\mathrm{lr}}$ to encode LR frames. 
Note that $F_{\mathrm{hr}}$ and $F_{\mathrm{lr}}$ are learnable to ensure both HR and LR frames can be encoded in the same space before feeding into Hybrid Transformer.
We extract hybrid spatiotemporal feature $\mathcal{V}_\mathrm{hy}$ of segment $s_i$ as the output of $T_\mathrm{hy}$. In addition, we use the HR frame feature extracted by stage 3 of $F_\mathrm{hr}$ (denoting as $F^3_\mathrm{hr}$) as HR input of $T_\mathrm{hy}$:
\vspace{-1mm}
\begin{equation}
    \mathcal{V}_\mathrm{hy} = T_\mathrm{hy}(F_\mathrm{lr}(\boldsymbol{X}^{s_i}_{t-Nr}), ..., \phi(F^3_\mathrm{hr}(\boldsymbol{X}^{s_i}_{t})),...),
\vspace{-1mm}
\end{equation}
where $\phi$ is an interpolate operation to align feature size.
In $T_\mathrm{hy}$, we adopt Divided Space-Time Attention to encode spatiotemporal information similar to \cite{gberta_2021_ICML}. 
We extract detailed spatial feature $\mathcal{V}_{\mathrm{hr}}$ of segment $s_i$ as the output of $E_{\mathrm{hr}}$ by:
\vspace{-1mm}
\begin{equation}
    \mathcal{V}_\mathrm{hr} = D_\mathrm{hr}(F_\mathrm{hr}(\boldsymbol{X}^{s_i}_{t})),
\vspace{-1mm}
\end{equation}
To adapt the output of the HR encoder to the hybrid spatiotemporal feature $\mathcal{V}_\mathrm{hy}$, $D_\mathrm{hr}$ consists of a convolution layer to adjust the output feature channel, as well as a $2\times 2$ max-pooling layer for down-sampling.
The segment features is the fusion of $\mathcal{V}_\mathrm{hr}$ and $\mathcal{V}_\mathrm{hy}$ by a linear layer:
\vspace{-1mm}
\begin{equation}
    \mathcal{V} = \boldsymbol{Linear}([\mathcal{V}_\mathrm{hr},\mathcal{V}_\mathrm{hy}]).
\vspace{-1mm}
\end{equation}

\vspace{-1mm}
\subsection{Language Encoder and Multi-Modality Joint Embedding Learning}\label{subsec: transformer}
\vspace{-1mm}
For both language encoder and multi-modality joint embedding learning, we use self-attention to model the relationship of both uni-modality and multi-modality. 
More specifically, we adopt a 24-layer, 1024-dimensional Transformer, mirroring the BERT-large and initialize it with pre-trained BERT-large parameters. We use the first 12 layers as language-only Transformer and the last 12 layers as multi-modal Transformer. Language-only Transformer extracts language representation which is concatenated with video features of a segment as the input of multi-modal Transformer. We add learnable 1D and 2D position embedding to language and vision tokens, respectively. Such a modal-independent design has two advantages. 
Firstly, it enables to provide powerful embedding for a single-modal input in downstream tasks. For example, the vision-aware language-only embedding could be used for language-guided video generation tasks.
Secondly, the two-stream architecture improves the calculation efficiency of similarity between video and language to linear complexity in some specific downstream tasks, such as video-language retrieval.

\begin{table*}[!t]
\small
\centering
\subfloat[MRSVTT-QA test set.
\label{tab:main_msrvtt_qa}]{
\begin{tabular}{l c}
\hline
\multicolumn{1}{c}{Method} & Acc \\
\hline\hline
ST-VQA~\cite{jang2017tgif} & 30.9  \\
Co-Memory~\cite{gao2018motion} & 32.0 \\
AMU~\cite{xu2017video} & 32.5 \\
Heterogeneous Mem~\cite{fan2019heterogeneous} & 33.0 \\
HCRN~\cite{le2020hierarchical} & 35.6 \\
ClipBERT PT~\cite{lei2021less}  & 37.4 \\
\hline
Ours & \bf 40.0 \\
\hline
\end{tabular}
}
\hspace{5mm}
\subfloat[MRSVTT multiple-choice test. \label{tab:main_msrvtt_multiple_choice_test}]{
\begin{tabular}{l c }
\hline
\multicolumn{1}{c}{Method} & Acc \\
\hline\hline
CT-SAN~\cite{yu2017end}  & 66.4 \\
MLB~\cite{kim2016hadamard}  & 76.1 \\
JSFusion~\cite{yu2018jsfusion} & 83.4 \\
ActBERT PT~\cite{zhu2020actbert} & 85.7 \\
ClipBERT PT~\cite{lei2021less} &  88.2 \\
VideoClip PT~\cite{xu2021videoclip} & 92.1 \\
\hline
Ours & \bf 97.1 \\
\hline
\end{tabular}
}
\hspace{5mm}
\subfloat[TGIF-QA test set.
\label{tab:main_tgif_qa}]{
\begin{tabular}{l c  c  c}
\hline
\multicolumn{1}{c}{Method} & Action & Trans & Frame \\
\hline\hline
ST-VQA~\cite{jang2017tgif} & 60.8 & 67.1 & 49.3 \\
Co-Memory~\cite{gao2018motion} & 68.2 & 74.3 & 51.5 \\
PSAC~\cite{li2019beyond} & 70.4 & 76.9 & 55.7 \\
HCRN~\cite{le2020hierarchical} & 75.0 & 81.4 & 55.9 \\
QueST~\cite{jiang2020divide} & 75.9 & 81.0 &  59.7 \\
ClipBERT PT~\cite{lei2021less} &  82.8 &  87.8 & 60.3 \\
\hline
Ours & \textbf{84.3} & \textbf{90.0} & \textbf{60.5} \\
\hline
\end{tabular}
}
\vspace{-3mm}
\caption{
Comparison of HD-VILA with state-of-the-art methods on video question answering tasks. (a) Results of ST-VQA and Co-Memory are implemented by \cite{fan2019heterogeneous}. (b) Results of CT-SAN and MLB are implemented by \cite{yu2018jsfusion}.}
\label{tab:main_video_qa}
\vspace{-5mm}
\end{table*}

\vspace{-1mm}
\subsection{Pre-Training Tasks}\label{subsec:pre_tasks}
\vspace{-1mm}
We adopt two pre-training tasks in HD-VILA: video-language matching to enhance cross-modal matching and masked language modeling (MLM) to encourage the mapping between visual and language tokens in fine-grained level. In particular, since the matching between video and language is somewhat weak compared with the video description dataset, we apply contrastive video-language matching to take advantage of large data. 
\vspace{-5mm}
\paragraph{Contrastive Video-Language Matching}
To align the feature space of video and language, we use a contrastive loss to maximize the similarity of a video clip and a sentence. 
Specifically, we treat matched pairs in a batch as positives, and all other pairwise combinations as negatives:
\vspace{-3mm}
\begin{equation}
\begin{aligned}
\mathcal{L}_{v 2 t} &=-\frac{1}{B} \sum_{i=1}^{B} \log \frac{\exp \left(v_{i}^{\top} t_{i} / \tau\right)}{\sum_{j=1}^{B} \exp \left(v_{i}^{\top} t_{j} / \tau\right)} \\
\mathcal{L}_{t 2 v} &=-\frac{1}{B} \sum_{i=1}^{B} \log \frac{\exp \left(t_{i}^{\top} v_{i} / \tau\right)}{\sum_{j=1}^{B} \exp \left(t_{i}^{\top} v_{j} / \tau\right)},
\end{aligned}
\end{equation}
where $v_i$ and $t_j$ are the normalized embeddings of $i$-th video and $j$-th sentence in a batch of size $B$ and $\tau$ is the temperature. Video and sentence features are computed by our hybrid video encoder and language encoder. The mean of segment embeddings is used as the video-level embedding.

\vspace{-5mm}
\paragraph{Masked Language Modeling}
We adopt Masked Language Modeling (MLM) to better build the mapping between visual and language domain. MLM aims to predict the ground-truth labels of masked text tokens from the contextualized tokens:
\vspace{-3mm}
\begin{equation}
\mathcal{L}_{\mathrm{MLM}}=-\mathbb{E}_{(\mathcal{W}, \mathcal{V})} \log p\left(w_{i} \mid \mathcal{W}_{\backslash i}, \mathcal{V}\right),
\vspace{-1mm}
\end{equation}
where $\mathcal{W}$ denotes the text embedding token set, $\mathcal{V}$ denotes the visual token set, and $w_i$ denotes the masked token. $(\mathcal{W}, \mathcal{V})$ is sampled from the distribution of text-video pairs. 
We adopt the same masking strategy as in BERT and use an MLP as the MLM head to output logits over vocabulary, which is then computed as the negative log-likelihood loss for the masked token. We aggregate the logits of different segments to derive a consensus, so that MLM is able to be calculated in video-level as we adopt in the approach.
\vspace{-2mm}
\section{Experiments}
\vspace{-1mm}

In this section, we conduct extensive experiments to evaluate the proposed HD-VILA pre-training model. 

\vspace{-1mm}
\subsection{Pre-training Details}
\vspace{-1mm}
Inspired by the idea of ``align before fuse''\cite{Li2021AlignBF}, we adopt a two-stage fashion for pre-training on HD-VILA-100M dataset.  In the first stage, we perform the contrastive video-language matching task to learn cross-modality alignment. In the second stage, MLM task is performed to facilitate cross-modal understanding. For video encoder, we use ResNet-50 for $F_{hr}$ and $F_{lr}$, and a 4-layer Transformer with 16 heads and 1024 hidden size for $T_{hy}$. We empirically divide a video clip into two segments and sample seven frames for each segment. In this setting, the two segments can cover about 6s video content, which are adequate to model the video clips in our dataset. Besides, we randomly crop $640 \times 1024$ areas for the middle HR frames, and select aligned $160 \times 256$ areas for LR neighboring frames. The size of resultant feature map before feeding into the multimodal Transformer is $10\times16$. For language, we follow BERT~\cite{Devlin2018} to adopt the WordPiece tokenizer to split a sentence into word tokens with a max length of 50. 

In pre-training, we use AdamW optimizer~\cite{loshchilov2017adamw} with an initial learning rate of 5e-5 and a fixed weight decay of 1e-3. We also employ a linear decay learning rate schedule with a warm-up strategy. We train our model with 128 NVIDIA Tesla V100 GPUs for stage one and 32 for stage two. The batch size is set to 1024 and the contrastive similarity is calculated on gathered features from all GPUs. We train 7 epochs for stage one and 4 epochs for stage two empirically. We freeze the encoders during the second stage and keep the same batch size for both stages. In downstream tasks, we keep the same model configuration if not otherwise specified. We exclude the YouTube Ids in the downstream tasks from our collected HD-VILA-100M during pre-training.

\begin{table}[]
    \small
    \centering
    \begin{tabular}{l c c c c c} 
    \hline
    Method  & R@1 $\uparrow$ & R@5 $\uparrow$ & R@10 $\uparrow$ & MedR $\downarrow$ \\\hline\hline
    HowTo100M~\cite{miech2019howto100m}  &  14.9& 40.2& 52.8& 9.0  \\ 
    CE~\cite{liu2019use}  & 20.9 & 48.8 &62.4 &6.0  \\
    DECEMBERT~\cite{tang2021decembert} & 17.5 & 44.3& 58.6& 9.0  \\
    HERO~\cite{li2020hero} & 16.8 & 43.4& 57.7& -  \\
    ClipBERT~\cite{lei2021less} & 22.0 & 46.8& 59.9& 6.0  \\
    VLM~\cite{xu2021vlm} & 28.1 & 55.5 & 67.4 &4.0  \\
    \textcolor{lightgray}{MMT~\cite{gabeur2020multi}}  & \textcolor{lightgray}{26.6} &\textcolor{lightgray}{57.1} &\textcolor{lightgray}{69.6} &\textcolor{lightgray}{4.0}  \\
    \textcolor{lightgray}{Support Set~\cite{patrick2021supportset}}  & \textcolor{lightgray}{30.1} & \textcolor{lightgray}{58.5}& \textcolor{lightgray}{69.3}& \textcolor{lightgray}{3.0} \\
    VideoCLIP~\cite{xu2021videoclip}  & 30.9 & 55.4& 66.8& - \\\hline
    Ours  & \textbf{35.6} & \textbf{65.3}& \textbf{78.0} & \bf 3.0 \\\hline\hline
    Zero-shot \\\hline
    HT MIL-NCE~\cite{miech2020end}  & 9.9 & 24.0 & 32.4 & 29.5  \\ 
    \textcolor{lightgray}{Support Set~\cite{patrick2021supportset}}  & \textcolor{lightgray}{8.7} & \textcolor{lightgray}{23.0} & \textcolor{lightgray}{31.1} & \textcolor{lightgray}{31.0} \\
    VideoCLIP~\cite{xu2021videoclip}  & 10.4 & 22.2 & 30.0 & -\\\hline
    Ours  & \textbf{14.6} & \textbf{34.4}& \textbf{44.1}& \bf 15.0 \\\hline
    \end{tabular}
    \vspace{-3mm}
    \caption{Comparison of text-to-video retrieval in MSR-VTT~\cite{xu2016msr}. We gray out some lines to highlight fair comparisons with traditional retrieval models and general pre-training models. This mark is also applicable to Table \ref{tab:retrieval-lsmdc}, \ref{tab:retrieval-activitynet}.}
    \label{tab:retrieval-msrvtt}
    \vspace{-6mm}
\end{table}

\vspace{-1mm}
\subsection{Video Question and Answering}
\vspace{-1mm}
\paragraph{Datasets} 

\textbf{(a) MSRVTT-QA}~\cite{xu2017video} is created based on video and captions in MSR-VTT \cite{xu2016msr}. Given a question in a complete sentence, the model selects an answer from a pre-defined set. \textbf{(b) MSRVTT multiple-choice test}~\cite{yu2018jsfusion} is a multiple-choice task with videos as queries, and captions as answers. Each video contains five candidate captions, with only one positive match. \textbf{(c) TGIF-QA}~\cite{jang2017tgif} is build on GIF videos. We experiment with three TGIF-QA tasks:  
\textit{Action} is defined as a multiple-choice task to identify an action that has been repeated in a video.
\textit{Transition} aims to identify the state before or after another state. 
\textit{FrameQA} is about open-ended questions about the given video. The task objective is identical to MSRVTT-QA. 
More details are in the supplementary materials.

\vspace{-5mm}
\paragraph{Implementation  Details} 
For TGIF Action and Transition, we respectively concatenate five candidate answers with the question into five sequences. On top of the [CLS] token of the question, we train a two-layer MLP to predict the confidence of the five candidates with cross-entropy loss. For MSRVTT-QA and TGIF Frame, we encode the answers in a one-hot fashion, and train 2-layer MLP classifier over all answer candidates with a cross-entropy loss on-top of the [CLS] token of the question. For MSRVTT Multiple-choice, we directly choose the answer with the highest similarity.
We set the max batch size to fine-tune on 8 V100 32G GPUs. More details are in the supplementary materials.

\vspace{-5mm}
\paragraph{Results}
In Table \ref{tab:main_video_qa}, the results of HD-VILA on video QA show that our model outperforms existing methods on five tasks in all the three datasets. On MSRVTT-QA and MSRVTT multiple-choice tests, we achieve \textbf{2.6} and \textbf{5.0} absolute improvement over SOTA methods. On TGIF-QA dataset, we have \textbf{1.5}, \textbf{2.2} and \textbf{0.2} absolute improvements on \textit{Action}, \textit{Trans} and \textit{Frame} tasks. The limited gain of \textit{Frame} is due to that \textit{Frame} focuses on single frame while hindering the advantage of our hybrid image sequence. Among all the compared methods, ClipBERT~\cite{lei2021less} and ActBERT~\cite{zhu2020actbert} are pre-training models. We can see that pre-training with more data will marginally improve the performance. Compared with ClipBERT which is pre-trained on image-language dataset, videos provide richer information. Note that the language used in ClipBERT pre-training is more closer to downstream tasks in both content and length while the language in HD-VILA-100M has domain gap with TGIF and MSR-VTT languages. This further indicates the generalization of the video representation learned by our HD-VILA.

\begin{table}[]
    \small
    \centering
    \begin{tabular}{l c c c c c} 
    \hline
    Method  & R@1 $\uparrow$ & R@5 $\uparrow$ & R@10 $\uparrow$ & MedR $\downarrow$ \\\hline\hline
    HERO~\cite{li2020hero} & 2.1 & - & 11.4& -  \\ 
    S2VT~\cite{venugopalan2014translating}  &  11.9& 33.6& - & 13.0  \\ 
    FSE~\cite{zhang2018cross}  &  13.9& 36.0& - & 11.0  \\ 
    CE~\cite{liu2019use}  & 16.1 & 41.1 & - &8.3  \\
    ClipBERT~\cite{lei2021less} & 20.4 & 48.0& 60.8& 6.0  \\
    \hline
    Ours & \textbf{28.8} & \textbf{57.4}& \textbf{69.1} & \textbf{4.0} \\\hline
    \end{tabular}
    \vspace{-3mm}
    \caption{Comparison of text-to-video retrieval on DiDeMo~\cite{anne2017localizing}. }
    \label{tab:retrieval-didemo}
    \vspace{-2mm}
\end{table}

\begin{table}[]
    \small
    \centering
    \begin{tabular}{l c c c c c} 
    \hline
    Method  & R@1 $\uparrow$ & R@5 $\uparrow$ & R@10 $\uparrow$ & MedR $\downarrow$ \\\hline\hline
    JSFusion~\cite{yu2018jsfusion}  &  9.1 & 21.2 & 34.1 & 36.0  \\ 
    MEE~\cite{miech2018learning}  &  9.3 & 25.1 & 33.4 & 27.0  \\ 
    CE~\cite{liu2019use}  & 11.2 & 26.9 & 34.8 & 25.3  \\
    \textcolor{lightgray}{MMT~\cite{gabeur2020multi}}  & \textcolor{lightgray}{12.9} & \textcolor{lightgray}{29.9} & \textcolor{lightgray}{40.1} & \textcolor{lightgray}{19.3}  \\
    \hline
    Ours  & \textbf{17.4} & \textbf{34.1}&  \textbf{44.1} & \textbf{15.0} \\ \hline
    \end{tabular}
    \vspace{-3mm}
    \caption{Comparison of text-to-video retrieval on LSMDC~\cite{Rohrbach2016MovieD}.}
    \label{tab:retrieval-lsmdc}
    \vspace{-2mm}
\end{table}

\begin{table}[]
    \small
    \centering
    \begin{tabular}{l c c c c c} 
    \hline
    Method  & R@1 $\uparrow$ & R@5 $\uparrow$ & R@50 $\uparrow$ & MedR $\downarrow$ \\\hline\hline
    FSE~\cite{zhang2018cross}  &  18.2& 44.8 & 89.1 & 7.0  \\ 
    CE~\cite{liu2019use}  & 18.2 & 47.7 & 91.4 & 6.0  \\
    HSE~\cite{zhang2018cross}  & 20.5 & 49.3 & - & -  \\
    ClipBERT~\cite{lei2021less} & 21.3 & 49.0& -& 6.0  \\
    \textcolor{lightgray}{MMT~\cite{gabeur2020multi}}  & \textcolor{lightgray}{28.7} & \textcolor{lightgray}{61.4} & \textcolor{lightgray}{94.5} & \textcolor{lightgray}{3.3}  \\
    \textcolor{lightgray}{Support Set~\cite{patrick2021supportset}}  & \textcolor{lightgray}{29.2} & \textcolor{lightgray}{61.6} & \textcolor{lightgray}{94.7} & \textcolor{lightgray}{3.0}  \\
    \hline
    Ours  & \textbf{28.5} & \textbf{57.4}& \textbf{94.0} & \textbf{4.0} \\\hline
    \end{tabular}
    \vspace{-3mm}
    \caption{Comparison of text-to-video retrieval on ActivityNet~\cite{Krishna2017actnetcaption}.}
    \label{tab:retrieval-activitynet}
    \vspace{-6mm}
\end{table}

\vspace{-1mm}
\subsection{Video-Text Retrieval}
\vspace{-1mm}

\paragraph{Datasets} We conduct video-text retrieval experiments on four datasets. \textbf{(a) MSR-VTT}~\cite{xu2016msr} contains 10K YouTube videos with 200K descriptions. We follow previous works~\cite{yu2018jsfusion,liu2019use}, training models on 9K videos, and reporting results on the 1K-A test set. \textbf{(b) DiDeMo}~\cite{anne2017localizing} consists of 10K Flickr videos annotated with
40K sentences. We follow~\cite{liu2019use, zhang2018cross} to evaluate paragraph-to-video retrieval, where all descriptions for a video are concatenated to form a single query. \textbf{(c) LSMDC}~\cite{Rohrbach2016MovieD} consists of 118,081 video clips sourced from 202 movies. Each video has a caption. Evaluation is conducted on a test set of 1,000 videos from movies disjoint from the train and validation sets. \textbf{(d) ActivityNet Captions}~\cite{Krishna2017actnetcaption} contains 20K YouTube videos annotated with 100K sentences. We follow the paragraph-to-video retrieval protocols \cite{zhang2018cross,liu2019use} training on 10K videos and reporting results on the val1 set with 4.9K videos.

\vspace{-5mm}
\paragraph{Implementation Details} We adjust the number of sampled segments and frames according to the average time of videos for each dataset. We adopt stage one model and the same training methods and objective for fine-tuning. We resize HR frame of each segment to 720p and LR frames to 180p. More details are in the supplementary materials.


\begin{figure}[t]
 \centering
 \includegraphics[width=\linewidth,trim=15 0 10 0]{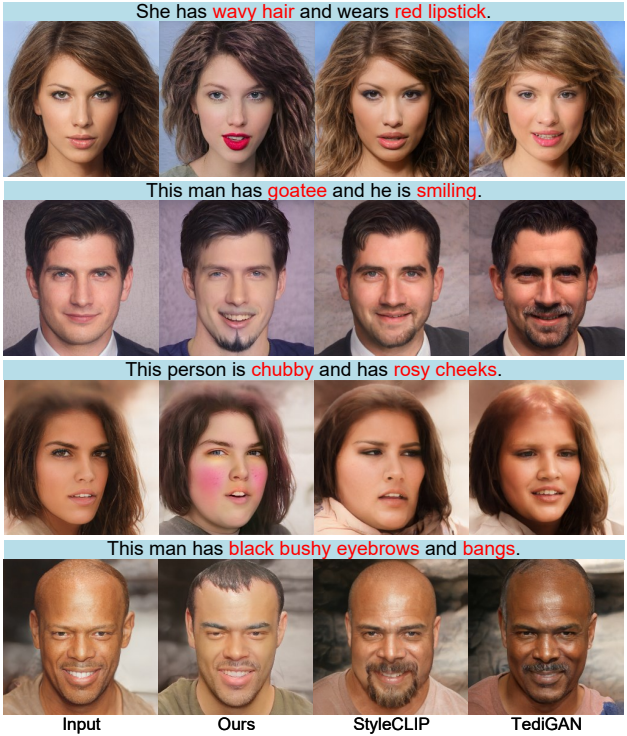}
 \vspace{-8mm}
 \caption{Text-guided manipulation compared with StyleCLIP \cite{patashnik2021styleclip} and TediGAN \cite{xia2021tedigan}.  Our model is able to handle complex descriptions and edit the inputs according to the target attributes (highlighted in {\color{red}red}) better. All the inputs are of \(1024\times 1024\) size.}
 \label{fig:man}
 \vspace{-5mm}
\end{figure}

\vspace{-5mm}
\paragraph{Results} Table \ref{tab:retrieval-msrvtt}, \ref{tab:retrieval-didemo}, \ref{tab:retrieval-lsmdc}, \ref{tab:retrieval-activitynet} show the text-to-video retrieval results of HD-VILA on four datasets. For MSR-VTT, we outperform the previous works by large margins in both zero-shot and fine-tuning settings. In particular, compared with VideoCLIP \cite{xu2021videoclip}, we have \textbf{40.4\%} relatively gains of R@1 in zero-shot setting, which shows the generalization ability of our pre-trained feature.
In LSMDC, we further obtain much larger relative gains with \textbf{55.4\%} under fair comparison. This comes from smaller domain gap between movie videos in LSMDC and our HD-VILA-100M compared with HowTo100M in two aspects: semantic (both open domains) and resolution (both high-resolution). On DiDeMo and ActivityNet, our model also achieves better performance. The videos in these two datasets are diversified in both scale and category, and are much longer. The results shows that our model pre-trained on HD-VILA-100M with longer videos and richer semantics shows better capacity for temporal understanding. Note that there are also pre-training models that are specifically designed for video-text retrieval task by improving noise contrastive learning like SupportSet~\cite{patrick2021supportset}, or use more features other than vision and motion like MMT~\cite{gabeur2020multi}. To make fair comparison, we gray them out in tables.

\begin{figure}[t]
 \centering
 \includegraphics[width=\linewidth,trim=15 0 10 0]{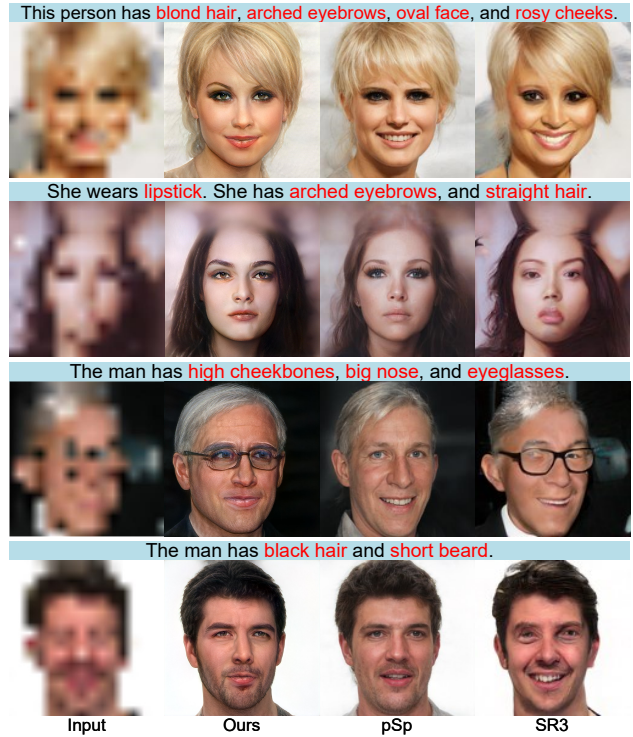}
 \vspace{-8mm}
 \caption{Text-guided super-resolution compared with pSp \cite{richardson2021psp} and SR3 \cite{saharia2021sr3}. Our model is able to reconstruct more accurate target attributes with descriptions (\textit{e.g.}, eyeglasses in the third case). All inputs are upsampled from \(16\times 16\) to \(1024\times 1024\).}
 \label{fig:sr}
 \vspace{-5mm}
\end{figure}

\vspace{-1mm}
\subsection{Text-to-Visual Generation}
\vspace{-1mm}

Recent studies like StyleCLIP\cite{patashnik2021styleclip} and TediGAN \cite{xia2021tedigan} propose to leverage cross-modal pre-training power to facilitate language-guided generation tasks, and have obtained some promising results. As shown in their work, the quality of visual generation results can reflect the quality of cross-modality embedding. Hence, in this section,  we will specify how our pre-trained model can achieve this task, and verify our learned embedding by showing higher-quality visualized results compared with SOTA models.  

\vspace{-5mm}
\paragraph{Datasets} To conduct this research, we introduce the first \textbf{F}ace-\textbf{D}escription-\textbf{V}ideo \textbf{D}ateset (\textbf{FDVD}). The dataset consists of 613 high-resolution (\(1024\times 1024\)) videos, resulting in 74,803 frames of human faces. The videos are collected from Ryerson audio-visual dataset \cite{livingstone2018ryerson}. We generate ten different text descriptions for each video following previous works \cite{xia2021tedigan}. To increase the diversity of human faces, we also leverage Multi-modal CelebA-HQ \cite{xia2021tedigan} for training. 

\vspace{-5mm}
\paragraph{Implementation Details} 
We follow previous works \cite{xia2021tedigan,patashnik2021styleclip} to leverage a well pre-trained StyleGAN \cite{karras2019stylegan,zeng2021improving,zheng2020learning} as our generator, due to its superior performance. 
In practice, we learn several linear layers to map the vision and text embedding in HD-VILA to the latent codes \(w^+\) used in StyleGAN. Then, images can be generated by the latent codes. To ensure the visual quality, identity preservation, and matching with descriptions of the generated results, we carefully choose 
a set of losses for optimization.
More details are in the supplementary materials.

\vspace{-5mm}
\paragraph{Text-to-Visual Editing} We compare our model with the recent state-of-the-art text-guided editing models, StyleCLIP \cite{patashnik2021styleclip} and TediGAN \cite{xia2021tedigan} in Figure \ref{fig:man}. The results show that our model is able to edit the target attributes of inputs according to text descriptions. For example, in the first case in Figure \ref{fig:man}, our model turns the hair to wavy hair and also wears lipstick on the lips, where StyleCLIP and TediGAN fail to wear lipstick on the face. Some video cases will be presented in supplementary materials.

\vspace{-5mm}
\paragraph{Text-to-Visual Super-Resolution} 
We further compare our model with SOTA super-resolution methods SR3 \cite{saharia2021sr3} and pSp \cite{richardson2021psp}. We generate $1024\times 1024$ images from their $16\times 16$ LR counterparts. Note that this task is extremely challenging due to such low-resolution inputs. As shown in the second case of Figure \ref{fig:sr}, SR3 \cite{saharia2021sr3} and pSp \cite{richardson2021psp} can not reconstruct high-quality faces by only using visual information. Compared with them, our model is able to accurately reconstruct the lipstick and the straight hair with the help of text description, thanks to the pre-trained models. 
\subsection{Ablation Studies}
\vspace{-1mm}
In this section, we conduct ablation studies to further verify the effectiveness of the new HD-VILA-100M dataset, and the proposed hybrid video encoder. 

\paragraph{(1) Diversity of HD-VILA-100M.} We sample two video subsets from HD-VILA-100M with two million clip-text pairs for each. One subset only includes ``HowTo'' type, while the other consists of diversified and balanced categories sampled from the full dataset. As shown in Table~\ref{tab:ablation2}, compared with the ``HowTo'' dataset with limited semantics, our diversified pre-training dataset (indicated as ``Ours-720p'') helps to achieve higher performance in the MSR-VTT retrieval task, with relative \textbf{66.7\%} R@1 gains. We choose MSR-VTT zero-shot retrieval task for this ablation study, as it is the most widely-used evaluation task in video-language pre-training. 
We also make fair comparison with HowTo100M~\cite{miech2019howto100m}. we have tried our best to collect HowTo100M at 720p, in which 69\% videos are originally at 720p, and 31\% are at 240p (w/o HR source) and upsampled to 720p by applying the most commonly used bicubic interpolation. We select MSR-VTT retrieval which is the most widely-used benchmark for pre-training evaluation. We report the comparison in Table~\ref{tab:rebut_1}. We compare pre-training on two datasets for the same steps (145K) and fine-tuning with the same setting. HD-VILA-100M pre-trained model surpasses HowTo100M by a large margin. This shows the advantage of HD-VILA-100M.

\vspace{-3mm}
\paragraph{(2) High-resolution of HD-VILA-100M.} We downsample ``Ours-720p'' subset into lower resolutions (``Ours-360p''), and observed a significant drop with \textbf{29.1\%} relative decreases of R@1. Such evaluations demonstrate the superiority of the diversified categories and higher resolution of the proposed dataset. 

\vspace{-3mm}
\paragraph{(3) Numbers of HR/LR frames.} As the number of high/low-resolution frames used for video modeling often plays a key role in video pre-training, we adjust frame numbers and fine-tune the pre-training model in different settings. As shown in  Table~\ref{tab:ablation1}, high-resolution frames lead to significant increases compared with the setting only using low-resolution inputs. In particular, the setting of 1-HR \& 10-LR achieves the best performance, compared with 0-HR \& 10-LR (``0'' indicates that one branch is removed), and  1-HR \& 0-LR, which demonstrates the rationality of jointly modeling spatial and temporal features in our approach. 

\begin{table}[]
\small
    \centering
    \begin{tabular}{l l r r r r} 
    \hline
    Type & Size & R@1 $\uparrow$ & R@5 $\uparrow$ & R@10 $\uparrow$ & MedR $\downarrow$ \\\hline\hline
    HowTo & 720p  & 3.3& 8.2& 13.5 & 113.0 \\ 
    Ours & 360p   & 3.9& 11.0& 18.3 & 67.0\\ 
    Ours & 720p*   & 4.5& 13.0& 20.2 & 62.0\\
    Ours & 720p   & \bf 5.5& \bf 13.1& \bf 20.5 & \bf 58.0 \\
    \hline
    \end{tabular}
    \vspace{-3mm}
    \caption{Ablation study on two subsets of pre-training data. We report results of zero-shot MSR-VTT retrieval. 720p* indicates bi-cubic upsampled frames (360p to 720p).}
    \label{tab:ablation2}
    \vspace{-2mm}
\end{table}

\begin{table}[]
\setlength\tabcolsep{4pt}
    \centering
    \begin{tabular}{l r r r r} 
    \hline
    Dataset & R@1 $\uparrow$ & R@5 $\uparrow$ & R@10 $\uparrow$ & MedR $\downarrow$ \\\hline\hline
    HowTo100M  &19.6 &	49.0 &	61.9  &	6.0  \\ 
    Ours  & \bf 30.0  & \bf 58.1  & \bf 72.3   & \bf 4.0 \\
    \hline
    \end{tabular}
    \vspace{-3mm}
    \caption{Comparison of pre-training datasets on MSR-VTT retrieval with the same steps.}
    \label{tab:rebut_1}
    \vspace{-2mm}
\end{table}

\begin{table}[]
\small
    \centering
    \begin{tabular}{c c c c c c} 
    \hline
    \#HR & \#LR & R@1 $\uparrow$ & R@5 $\uparrow$ & R@10 $\uparrow$ & MedR $\downarrow$ \\\hline\hline
    1 & 0 & 16.3 & 40.0 & 53.3 & 9.0 \\ 
    0 & 10 & 26.7 & 57.0 & 69.5 & 4.0  \\
    1 & 6 & 33.0 & 64.4 & 76.2 & 3.0   \\
    1 & 10 & \bf 35.6 & \bf 65.3 & \bf 78.0 & \bf 3.0  \\
    1 & 14 & 33.7 & 64.1 & 76.2 & \bf 3.0 \\
    \hline
    \end{tabular}
    \vspace{-3mm}
    \caption{Ablation study on frame selection. We report results of MSR-VTT retrieval, where \#HR/\#LR are the numbers of high/low-resolution frames. }
    \label{tab:ablation1}
    \vspace{-6mm}
\end{table}

\section{Conclusion}
\vspace{-1mm}
In this paper, we propose to learn high-resolution and diversified video-language multi-modal representation by pre-training on large-scale video-language pairs. To empower pre-training, we introduce a new dataset \textbf{HD-VILA-100M} which is the largest high-resolution and diversified video-language dataset. 
To more efficiently employ the richer information in videos, we propose a novel pre-training model HD-VILA that learns spatiotemporal information using HR and LR frames as a \textbf{hybrid} image sequence with a hybrid Transformer. 
Experiments on \textbf{12} video-language understanding and text-to-visual generation tasks show the capability of HD-VILA-100M dataset and the effectiveness of our model.

\paragraph{Acknowledgement}
We would like to thank the insightful discussion, valuable suggestions and kind help from Prof. Jiebo Luo, Prof. Ruihua Song, Prof. Limin Wang, Houwen Peng, and Dongdong Chen.

{\small
\bibliographystyle{ieee_fullname}
\bibliography{main}
}
\appendix
\section{Author Contribution}
The first four authors contribute equal to this research project. 
Among them, Hongwei Xue is responsible for model design, implementation of pre-training model and downstream video QA tasks. Tiankai Hang helps the model design, environment building for distributed training, and apply the pre-trained model for downstream extreme text-guided super-resolution task. Yanhong Zeng is in charge of the text-to-visual generation part, including the creation of dataset (FDVD), design and implementation of text-to-visual editing. Yuchong Sun is responsible for collecting and processing HD-VILA-100M dataset, discussing model design and implementation of downstream video-text retrieval tasks.
Bei Liu, Huan Yang and Jianlong Fu oversee the whole project, including dataset collection and processing, pre-training model and downstream tasks design. Baining Guo provides valuable suggestions in paper organization and writing. 

\section{Limitation and Social Impact}
The proposed video-language dataset and pre-training model show the capacity and generalization of learned video-language representation which could benefit many applications of computer vision and natural language processing. Pre-training with large scale of data results in much computation resource. How to reduce the model size and computing effort becomes more essential for future research. In addition, the usage of user generated data might bring the risk of bias. We tackle this problem by balancing various video categories, yet the videos might contain biased content. Moreover, how to avoid malicious usage of visual generation technique for conscious attack is also critical. However, these concerns are general to the entire fields and are not amplified by this work.

\begin{figure*}
    \centering
    \begin{subfigure}[b]{0.33\textwidth}
        \centering
        \includegraphics[width=\linewidth]{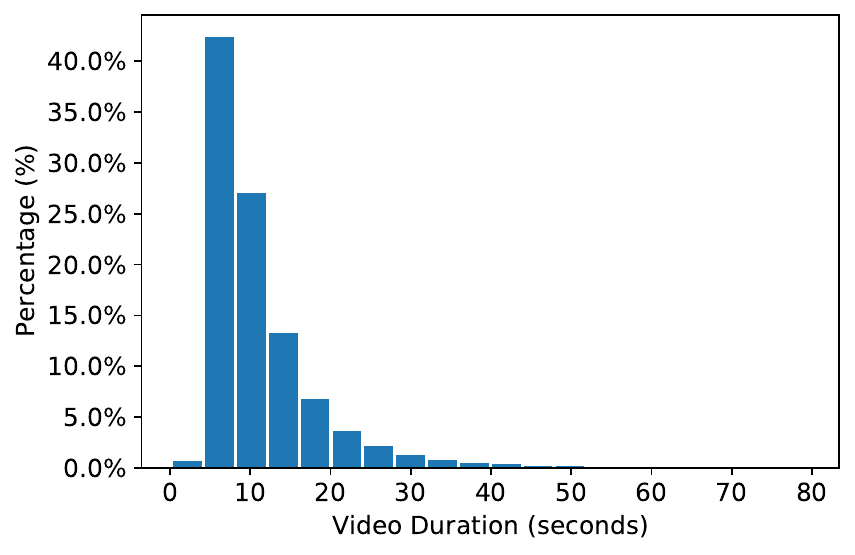}
        \caption{Distributions of video duration.}
        \label{fig:video_dur}
    \end{subfigure}
    \hfill
    \begin{subfigure}[b]{0.33\textwidth}
        \centering
        \includegraphics[width=\linewidth]{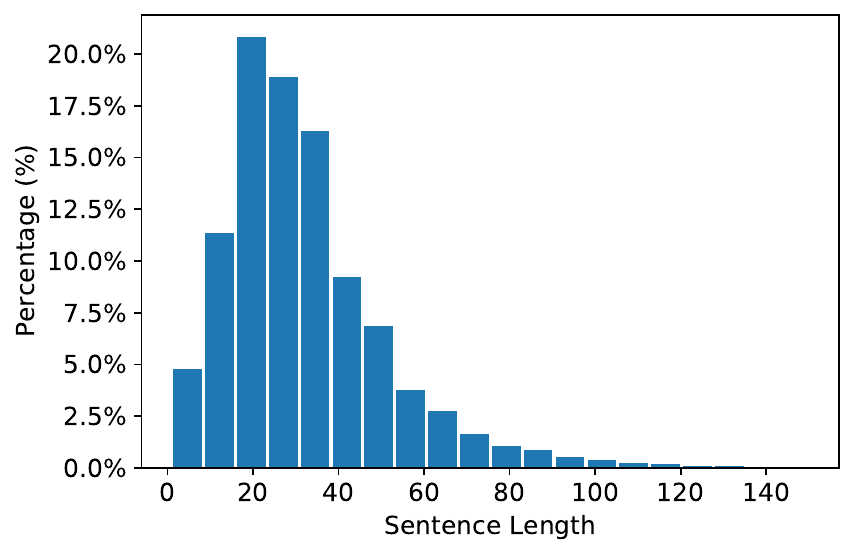}
        \caption{Distributions of sentence length.}
        \label{fig:num_words}
    \end{subfigure}
    \hfill
    \begin{subfigure}[b]{0.33\textwidth}
        \centering
        \includegraphics[width=\linewidth]{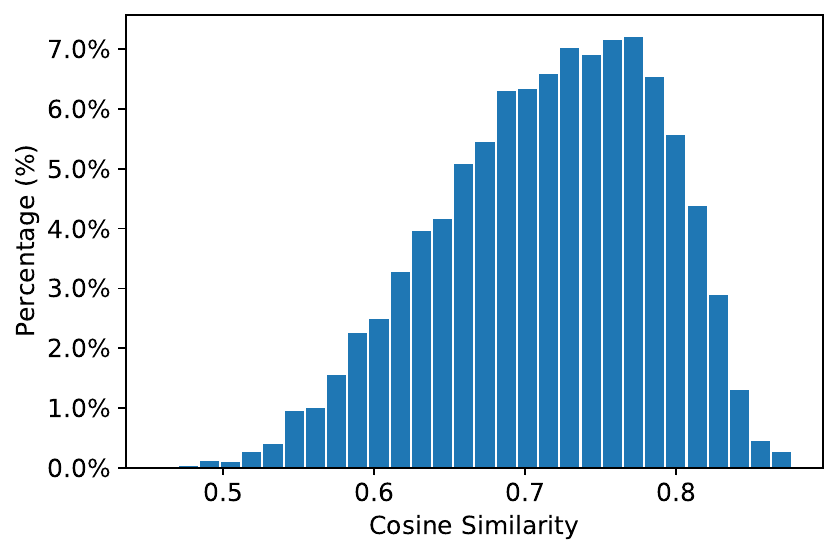}
        \caption{Video-text similarity distribution.}
        \label{fig:cos_sim}
    \end{subfigure}
\caption{More detailed statistics of HD-VILA-100M dataset.}
\end{figure*}

\section{HD-VILA-100M Dataset Details}

\subsection{Video Duration and Transcript Length}
We plot the histogram of video clip duration and transcript length in Figure \ref{fig:video_dur} and Figure \ref{fig:num_words}, respectively. 
From Figure \ref{fig:video_dur}, we can see that most video clips in our dataset is between 5s to 15s, with an average of 13.4s. From Figure \ref{fig:num_words}, most sentences in HD-VILA-100M are between 15 words to 50 words, with an average of 32.5 words.

\begin{table*}[t]
\setlength{\tabcolsep}{4pt}
\begin{center}
    \begin{tabular}{l r r r c r r r r }
    \toprule
    \multirow{2}{*}{\textbf{Dataset}}  & \multicolumn{3}{c}{\textbf{\# avg unique $n$-grams}} & ~ & \multicolumn{4}{c}{\textbf{\# avg POS tags}}\\
    \cmidrule(lr){2-4} \cmidrule(lr){6-9} 
     & 2-gram & 3-gram & 4-gram & ~ & noun & adj & adv & verb\\ 
    \midrule
    HowTo100M\cite{miech2019howto100m}     & 1.77 & 2.08 & 1.46 & ~ & 2.25 & 0.85 & 0.68 & 0.20\\
    HD-VILA-100M     & 4.18 & 13.08 & 20.89 & ~ & 6.63  & 1.88 & 2.07 & 5.09\\
    \bottomrule
    \end{tabular}
\end{center}
\caption{Statistics of average unique n-grams and POS tags. Our dataset has more unique n-grams and POS tags than HowTo100M\cite{miech2019howto100m}. The result indicates the transcriptions in HD-VILA-100M have richer and more diverse semantics.}
\label{table:howto-vila}
\end{table*} 

\subsection{Semantic Richness}
To analyze the semantic richness, we calculate the average unique n-grams and part-of-speech (POS) tags of transcriptions. We mainly compare them with HowTo100M~\cite{miech2019howto100m} dataset as shown in Table \ref{table:howto-vila}. From the result, we can find that the sentences in our dataset have more n-grams and POS tags, which indicates more richness and diversity of semantics in our HD-VILA-100M dataset.

\begin{figure*}
    \centering
    \includegraphics[width=0.85\linewidth]{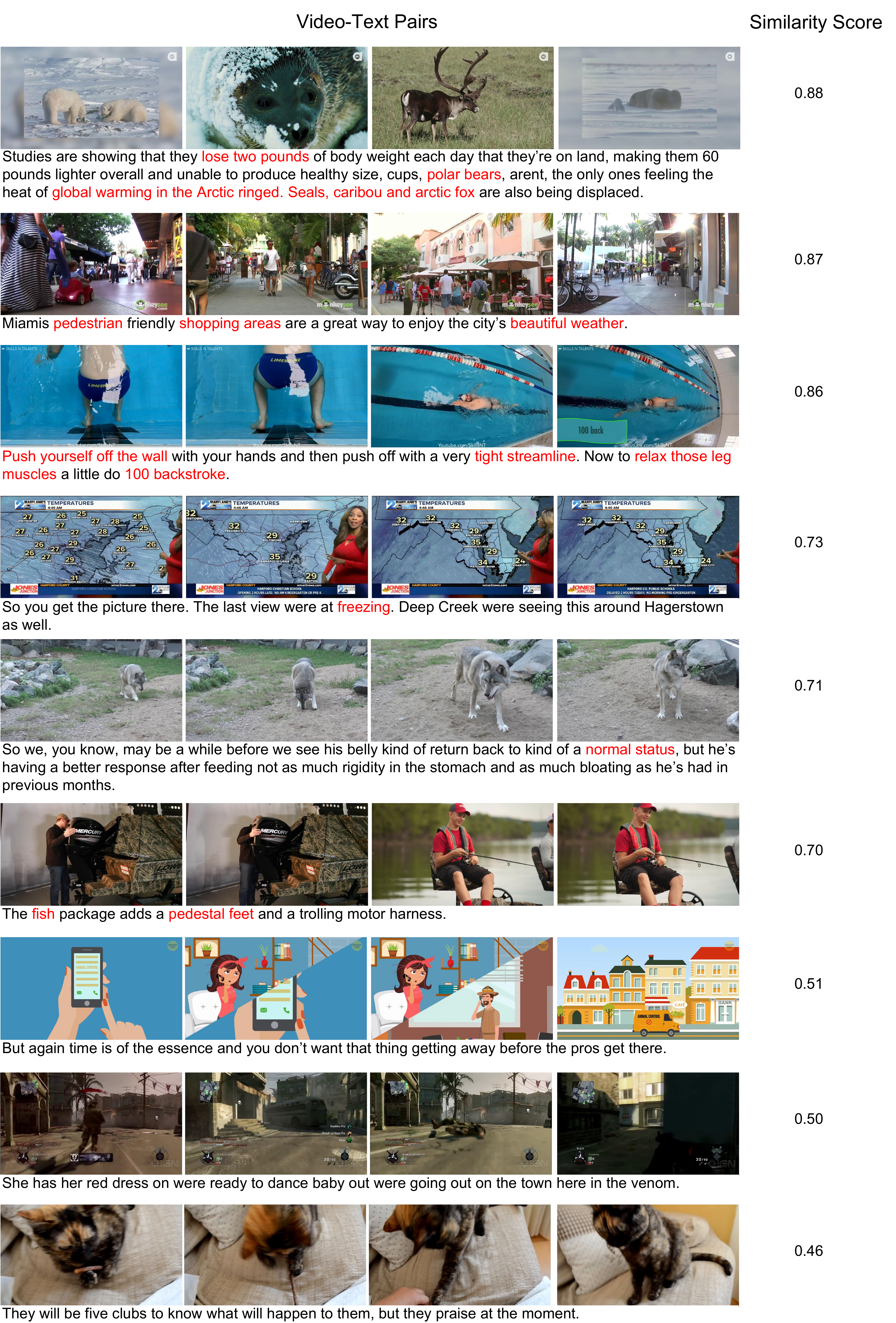}
    \caption{More examples of HD-VILA-100M with similarity scores calculated by HD-VILA. Relevant words are highlighted in red. [Best viewed in color.]}
    \label{fig:hdvila_case}
\end{figure*}

\subsection{More Examples of HD-VILA-100M Dataset}
Since we use transcripts as corresponding sentences for videos, the video-sentences are actually not all well aligned compared with video captioning datasets. Indeed, most of them are weakly related. We conduct an interesting experiment which uses our pre-trained model with the weakly aligned pairs to compute the similarity of these pairs. We show some examples with similarity scores in Figure \ref{fig:hdvila_case}. We can see that the pairs with higher score are well aligned. This indicates that, even with the weakly aligned video-transcript pairs, our pre-training model can learn a powerful embedding space between video and language.
The similarity score distribution of video and text pairs in HD-VILA-100M is shown in Figure \ref{fig:cos_sim}.

\section{Experiment Details}

\begin{figure*}
    \centering
    \includegraphics[width=\linewidth]{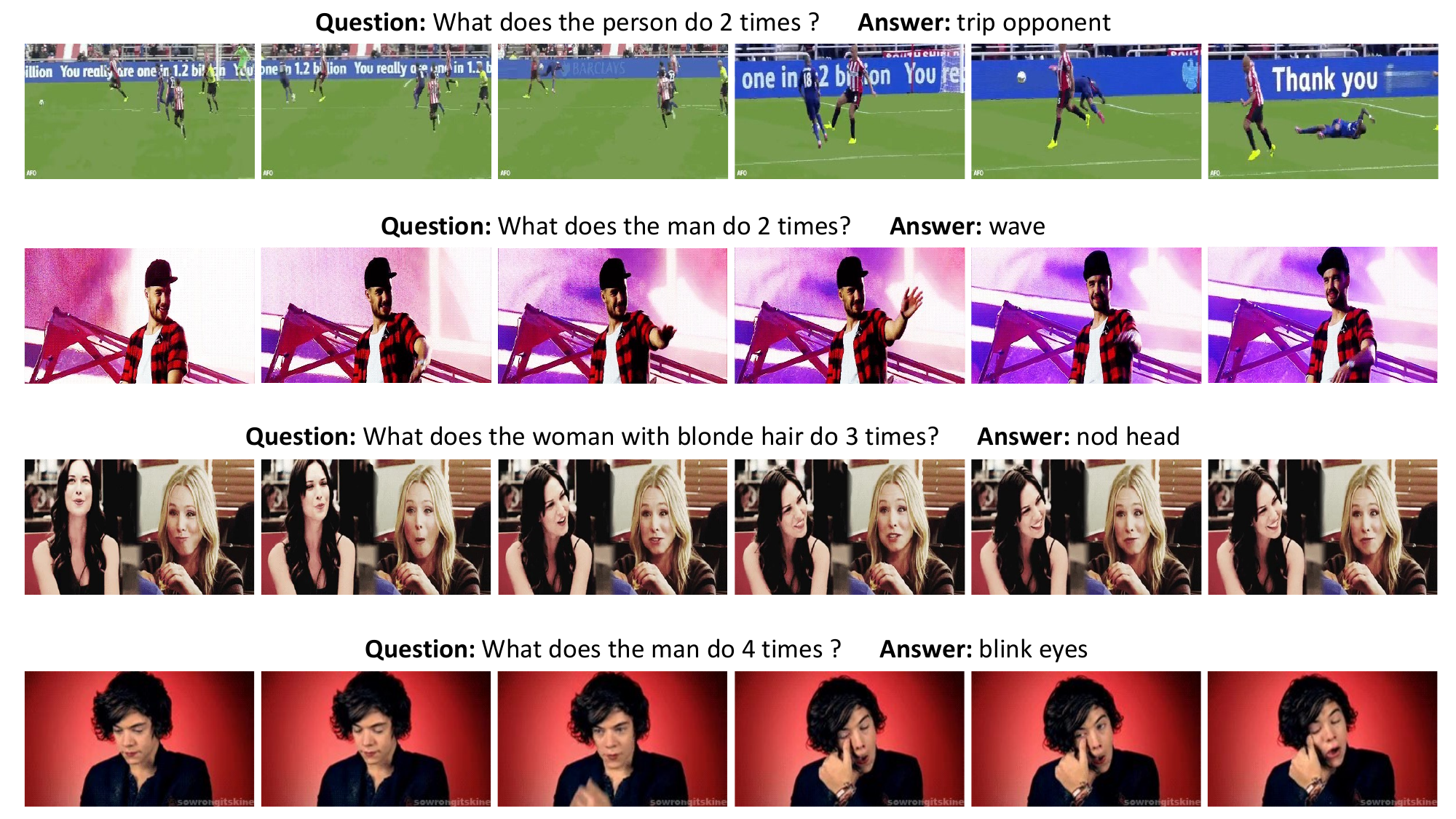}
    \caption{\textbf{Some examples for video QA task}. We take TGIF \textit{Action} for example to demonstrate our model's ability to learn temporal information from videos.}
    \label{fig:supp-qa}
\end{figure*}

\subsection{Video QA}
\paragraph{MSRVTT-QA.} 
MSRVTT-QA~\cite{xu2017video} is created based on video and captions in MSR-VTT~\cite{xu2016msr}, containing 10K videos and 243K open-ended questions. We follow the original work to use an answer vocabulary containing the most common 1.5K answers in the training and validation split as answer candidates. For each video, we randomly sample one segment for training and uniformly sample eight segments for testing. We resize HR frame of each segment to 720p and LR frames to 180p. In this task, we set \#HR as 1 and \#LR as 6.  We use AdamW for optimization, with an initial learning rate of 1e-5, weight decay of 0.3, and set learning rate warm-up over the first 10\% training steps followed by linear decay to 0. To alleviate over-fitting, we set dropout of Transformers to 0.1. We fine-tune our model on 8 NVIDIA Tesla V100 GPUs for 20 epochs with a batch size of 512. Gradient accumulation is applied to reach this batch size.

\paragraph{MSRVTT Multiple-Choice.}
MSRVTT multiple-choice test~\cite{yu2018jsfusion} is a multiple-choice task with videos as queries, and captions as answers. Each video contains five candidate captions, with only one positive match. The benchmark has 2,990 questions for the multiple-choice test. We directly inference our model trained on MSRVTT-Retrieval dataset to find the most positive match.

\paragraph{TGIF-QA.}
TGIF-QA~\cite{jang2017tgif} contains 165K QA pairs on 72K GIF videos. We experiment with three TGIF-QA tasks: \textit{Action}, \textit{Transition} and \textit{FrameQA}. We randomly sample one segment for training and uniformly sample eight segments for testing. Each segment contains 1 HR frame and 6 LR frames for \textit{Action} and \textit{Transition}, 10 LR frames for \textit{FrameQA}. Other settings are listed in Table \ref{tab:supp_tgif}. We use AdamW for optimization, and We fine-tune our model on 8 V100 GPUs, Gradient accumulation is applied to reach batch sizes listed in Table \ref{tab:supp_tgif}.

\begin{table}[hbt!]
\small
    \centering
    \begin{tabular}{l c c c} 
    \hline
    ~ & \textit{Action} & \textit{Transition}  & \textit{FrameQA} \\\hline\hline
    Epoch & 80  & 80 & 40 \\ 
    Batch Size & 384  & 384 & 448 \\ 
    Learning Rate & 5e-5 & 5e-5 & 4e-5\\ 
    Weight Decay & 0.05 & 0.05 & 0.3 \\
    Drop Out & 0.1 & 0.3 & 0.1 \\
    \hline
    \end{tabular}
    \vspace{-3mm}
    \caption{Details of training Video QA on TGIF dataset.}
    \label{tab:supp_tgif}
\end{table}

\subsection{Text-to-Video Retrieval}

Due to the various resolution for videos in downstream datasets, we resize HR frame of each segment to 720p and LR frames to 180p.  We adopt stage one model and the same training methods and objective for fine-tuning. We set the temperature to 0.08. We use learning rate warmup followed by multi-step learning rate decay. We adjust the number of sampled segments and frames according to the average time of videos for each dataset to cover about half of the video. For evaluation, we double the number of segments. More details for each tsak are given below

\paragraph{MSR-VTT.} MSR-VTT~\cite{xu2016msr} contains 10K YouTube videos with 200K descriptions. We follow previous works~\cite{yu2018jsfusion,liu2019use}, training models on 9K videos, and reporting results on the 1K-A test set. For zero-shot evaluation on low-resolution MSR-VTT videos, we uniformly sample 4 segments each with 11 frames. We crop a 224×320 patch for each frame and up-sample the middle frames by 4 times. In this setting, the sampled segments can nearly cover the videos on average. We remove the stop words in the text as~\cite{miech2020end}. We report the result of the last saved model of HD-VILA. When finetuning, we sample 2 segments for training and 4 segments for testing and each segment contains 11 frames. We use AdamW optimizer with an initial learning rate of 1e-5. We fine-tune the pre-trained model with 32 V 100 GPUs and the total batch size is 256.

\paragraph{DiDeMo.} DiDeMo~\cite{anne2017localizing} consists of 10K Flickr videos annotated with
40K sentences. We follow~\cite{liu2019use, zhang2018cross} to evaluate paragraph-to-video retrieval, where all descriptions for a video are concatenated to form a single query. When finetuning, we sample 4 segments for training and 8 segments for testing and each segment contains 11 frames. We use AdamW optimizer with an initial learning rate of 5e-6. We fine-tune the pre-trained model with 16 V 100 GPUs and the total batch size is 64.

\paragraph{LSMDC.} LSMDC~\cite{Rohrbach2016MovieD} consists of 118,081 video clips sourced from 202 movies. Each video has a caption. Evaluation is conducted on a test set of 1,000 videos. When finetuning, we sample 2 segments for training and 4 segments for testing and each segment contains 11 frames. We use AdamW optimizer with an initial learning rate of 5e-6. We fine-tune the pre-trained model with 8 V 100 GPUs and the total batch size is 64.

\paragraph{ActivityNet.} ActivityNet Captions~\cite{Krishna2017actnetcaption} contains 20K YouTube videos annotated with 100K sentences. We follow the paragraph-to-video retrieval protocols \cite{zhang2018cross,liu2019use} training on 10K videos and reporting results on the val1 set with 4.9K videos. When finetuning, we sample 4 segments for training and 8 segments for testing and each segment contains 13 frames. We use AdamW optimizer with an initial learning rate of 5e-6. We fine-tune the pre-trained model with 16 V 100 GPUs and the total batch size is 64.

\subsection{Text-to-Visual Generation}
In this section, we introduce more details about text-to-visual generation tasks as a supplement to Section 5.4 in the main paper. 
We introduce the details of model design in Section \ref{sec:mapper} and optimization objectives in Section \ref{sec:opt}, following the introduction of our collected dataset of video-description pairs of the human faces in Section \ref{sec:fdvd}. We provide more generation results and experimental analysis in Section \ref{sec:exp}.

\begin{figure}
    \centering
    \includegraphics[width=\linewidth]{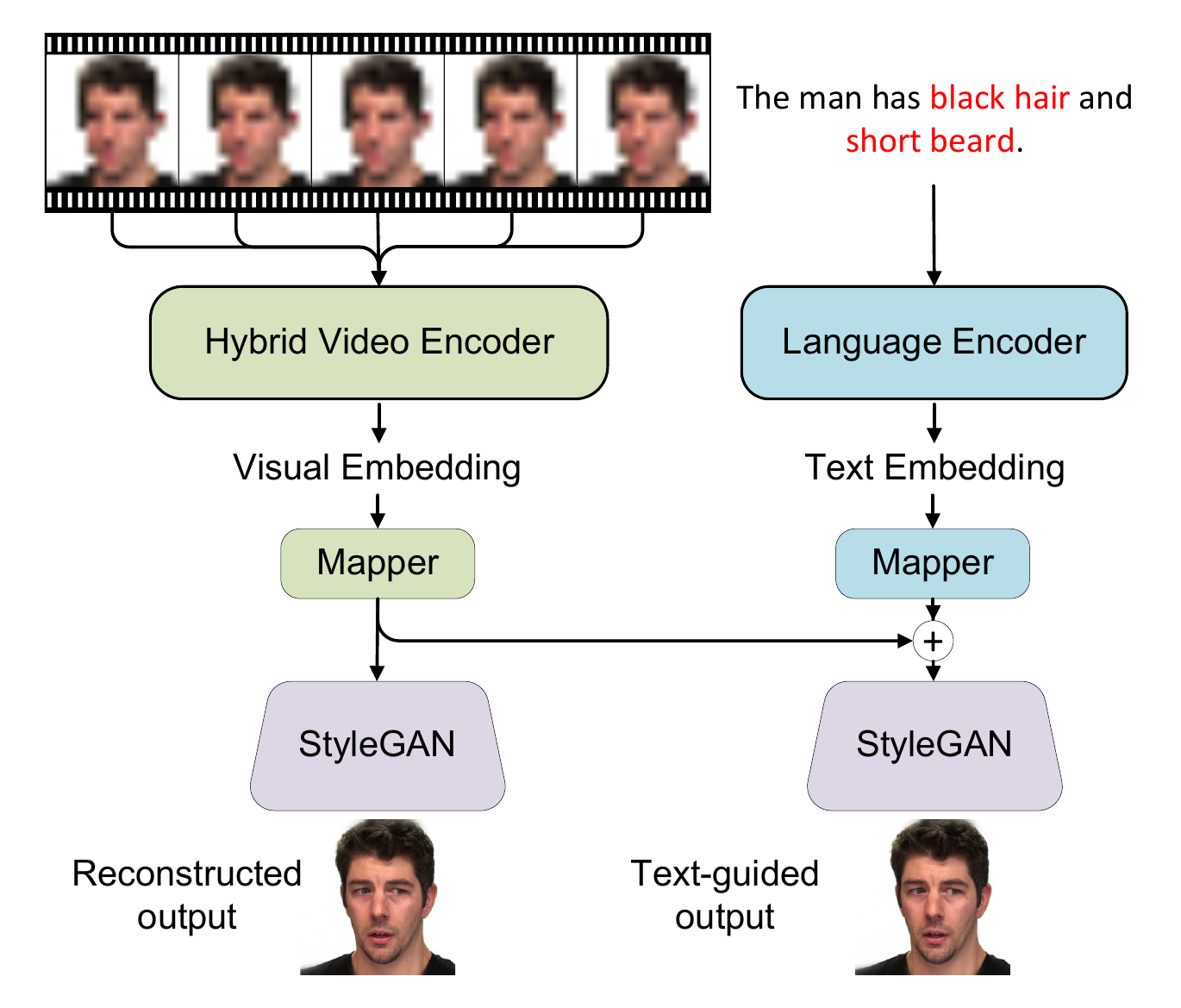}
    \caption{\textbf{Overview of our text-guided generation framework.}
    The framework consists of 1) two multi-modal encoders, 2) two mapper modules, and 3) a pre-trained StyleGAN \cite{karras2019stylegan}. First, the multi-modal encoders encode a video clip and a sentence to a visual and a text embedding, respectively. Second, the mapper modules map the embedding to the latent codes of StyleGAN. Finally, StyleGAN maps the latent codes w/ and w/o text information to images. See more details in Section \ref{sec:mapper}.}
    \label{fig:framework-gen}
\end{figure}

\subsubsection{Model Design}
\label{sec:mapper}
To achieve the text-to-visual generation tasks by our pre-trained model HD-VILA, we follow previous works to combine the cross-modality encoders of HD-VILA and a well pre-trained generation model, StyleGAN \cite{karras2019stylegan}, in our framework \cite{xia2021tedigan,patashnik2021styleclip}. The overview of our text-to-visual generation framework is shown in Figure~\ref{fig:framework-gen}. Specifically, our framework consists of three key components, including 1) two multi-modal (visual/text) encoders, 2) two visual/text mapper modules, and 3) a pre-trained StyleGAN. We introduce more details of each component as below. 

\paragraph{Multi-Modal Encoders} 
To deal with multi-modal inputs, we inherit the hybrid video encoder and the language encoder from our pre-trained model HD-VILA. Specifically, the hybrid video encoder takes as input a hybrid image sequence and outputs a visual embedding representing the input vision content. At the same time, the language encoder encodes the sentence into a text embedding that shares a joint embedding space with the visual embedding. Thanks to the large-scale pre-training on the proposed HD-VILA-100M dataset, the multi-modal encoders are able to provide vision-aware text embedding and text-aware vision embedding, which benefits downstream generation tasks. We denote the visual embedding and the text embedding as \(\mathbf{v}, \mathbf{t} \in \mathbb{R}^{1024}\) respectively.

\paragraph{Visual/Text Mappers} 
Since the output embedding \(\mathbf{v}, \mathbf{t} \in \mathbb{R}^{1024} \) of multi-modal encoders and the latent codes \(w^+ \in \mathbb{R}^{18\times 512}\) used for generation lie in different feature spaces, we build a visual mapper and a text mapper to bridge the gap between different feature spaces. Specifically, the mapping \(f\) is implemented using several layers MLP. It maps the embedding $\mathbf{v}, \mathbf{t}$ to $\mathbf{w}^+_{\mathbf{v}}, \mathbf{w}^+_{\mathbf{t}} \in \mathbb{R}^{18\times 512}$,
\begin{equation}
    \mathbf{w}^+_{\mathbf{v}} = f_v(\mathbf{v}), \ \ \ \ \ 
    \mathbf{w}^+_{\mathbf{t}} = f_t(\mathbf{t}),
\end{equation}
where $f_v, f_t$ denote the mapping functions.

\paragraph{Generator (StyleGAN)} 
Since StyleGAN has shown high-fidelity generation quality and impressive disentanglement property, we follow previous works to leverage a StyleGAN for generation \cite{xia2021tedigan,patashnik2021styleclip,karras2019stylegan}. Specifically, we incorporate a well pre-trained and fixed StyleGAN to generate images from the latent codes from mappers $\mathbf{w}^+_{\mathbf{v}}$ and $ \mathbf{w}^+_{\mathbf{t}}$.

In practice, the latent code \(\mathbf{w}^+_{\mathbf{v}}\) is optimized to reconstruct the high-quality middle frame in the input hybrid image sequence, while the latent code \(\mathbf{w}^+_{\mathbf{v}}\) is optimized to learn the editing directions according to the input sentences. Such a design enables keeping the information from visual inputs, as well as generating novel visual results according to the text inputs. We denote the reconstructed output and the text-guided output as:
\begin{equation}
    \mathbf{I}_{rec} = G(\mathbf{w}^+_{\mathbf{v}}), 
\end{equation}
\begin{equation}
    \mathbf{I}_{edit} = G(\mathbf{w}^+_{\mathbf{v}}+ \mathbf{w}^+_{\mathbf{t}}),
\end{equation}
where \(G\) denotes the synthesis network of StyleGAN. 

\subsubsection{Optimization Objectives}
\label{sec:opt}
To ensure per-pixel reconstruction accuracy, high-quality visual generation, identity preservation, and matching with the descriptions of the generated results, we carefully select a pixel-wise \(\mathcal{L}_2\) loss, a LPIPS loss \cite{zhang2018unreasonable}, an identity loss \cite{richardson2021psp}, and a text-visual matching loss as our optimization objectives following common practices \cite{tov2021designing,richardson2021psp,xia2021tedigan,patashnik2021styleclip}. Specifically, the pixel-wise \(\ell_2\) loss is denoted as: 
\begin{equation}
    \ell_2({\mathbf{I}},\hat{\mathbf{I}}) = || \mathbf{I} - \hat{\mathbf{I}}||_2, 
\end{equation}
where $\mathbf{I}$ denote the high-quality middle frame.
LPIPS is a deep metric that is able to reflect image quality similar to human perceptual \cite{zhang2018unreasonable}, and the LPIPS loss is denoted as: 
\begin{equation}
    \ell_{lpips}({\mathbf{I}},\hat{\mathbf{I}}) = ||\mathcal{F}(\mathbf{I}) - \mathcal{F}(\hat{\mathbf{I}}) ||_2,
\end{equation}
where $\mathcal{F}$ denotes the perceptual feature extractor.
We follow Richardson et al. to incorporate an identity recognition loss to measure the cosine similarity between the output image and its target \cite{richardson2021psp},
\begin{equation}
    \ell_{id}({\mathbf{I}},\hat{\mathbf{I}}) =  1 - \left \langle\mathcal{R}(\mathbf{I}), \mathcal{R}(\hat{\mathbf{I}})  \right \rangle,
\end{equation}
where $\mathcal{R}$ is a pre-trained network for face feature extractor, and \(\left \langle\cdot,\cdot  \right \rangle\) denotes cosine similarity calculation. To ensure the matching between the text-guided output and the input text, we follow StyleCLIP \cite{patashnik2021styleclip} to include a matching loss for optimization. In particular, the matching loss aims at minimizing the feature distance between the output image and the text,
\begin{equation}
    \ell_{clip}( \mathbf{T},\hat{\mathbf{I}}) =  1 - \left \langle\mathcal{C}(\mathbf{T}),\mathcal{C}(\hat{\mathbf{I}}) \right \rangle /\  \gamma ,
\end{equation}
where \(\mathcal{C}\) is a pre-trained image-text feature extractor,  \(\mathbf{T}\) denotes the text input, and \(\gamma\) is a constant value that normalize the similarity value to the range of [0,1]. In practice, we set the value of \(\gamma\) as 100.
The overall optimization objectives are concluded as:
\begin{math}
    \ell = \lambda_1 \cdot \ell_2(\mathbf{I},\mathbf{I}_{rec}) +  \lambda_2 \cdot \ell_{lpips}(\mathbf{I},\mathbf{I}_{rec}) +  \lambda_3 \cdot \ell_{id}(\mathbf{I},\mathbf{I}_{rec}) 
    + \lambda_4 \cdot \ell_2(\mathbf{I},\mathbf{I}_{edit}) + \lambda_5 \cdot \ell_{lpips}(\mathbf{I},\mathbf{I}_{edit}) + \lambda_6 \cdot \ell_{id}(\mathbf{I},\mathbf{I}_{edit})  
    + \lambda_7 \cdot \ell_{clip}( \mathbf{T},\mathbf{I}_{edit}).
\end{math}
\paragraph{Implementation Details}
We empirically set the loss weights for different generation tasks. For text-guided editing, we set $\lambda_1=1.0, \lambda_2 = 0.8,\lambda_3 = 0.1,\lambda_4 = 0.1,\lambda_5 = 0.1, \lambda_6=0.1, \lambda_7=1.0$. For text-guided super-resolution, we set $\lambda_1 = 0, \lambda_2 = 0, \lambda_3 = 0, \lambda_4 = 0.1, \lambda_5 = 0.8, \lambda_6 = 1.0,\lambda_7 = 0.1$. We use a fixed learning rate \(1e-5\) for the training of the multi-modal encoders, and a fixed learning rate \(1e-3\) for the visual/text mappers. We use Adam optimizer with $(\beta_1, \beta_2) = (0.9, 0.99)$ for training \cite{kingma2014adam}. We train the models for 200K iterations in total on 4 NVIDIA Tesla V100 GPUs.

\subsubsection{Face-Description-Video Dateset (FDVD)}
\label{sec:fdvd}
To demonstrate the effectiveness of our text-guided generation framework on videos, we collected a dataset of video-description pairs of the human faces, named \textbf{Face-Description-Video Dateset (FDVD)}. FDVD consists of 613 video-description pairs, resulting in 74,803 frames of human faces and 6,130 sentences in total. Specifically, each video-description pair consists of one high-resolution video (\(1024\times 1024\) spatial size) and ten different descriptive sentences. We introduce the collection process as below.

To generate high-quality videos of human faces, we collected videos from Ryerson audio-visual dataset \cite{livingstone2018ryerson}. For the pre-processing, we first use the facial landmark locations to select an appropriate crop region for the talking head, then we perform a high-quality up-sampling to obtain the final videos at \(1024\times 1024\) resolution following \cite{karras2018progressive}. 
To generate diverse descriptions for each video, we adopt a strategy of \textit{prediction-and-generation}. First, we use a facial attribute predictor \cite{liu2015faceattributes} to obtain a list of attributes for the videos. Then we follow previous best practices to use PCFG rule-based algorithm to generate descriptions from the given attributes \cite{xia2021tedigan,stap2020conditional}. Each description contains different subsets of the attributes to increase the diversity of descriptions. We will release the dataset for research purposes.

\begin{figure*}
    \centering
    \includegraphics[width=0.8\linewidth]{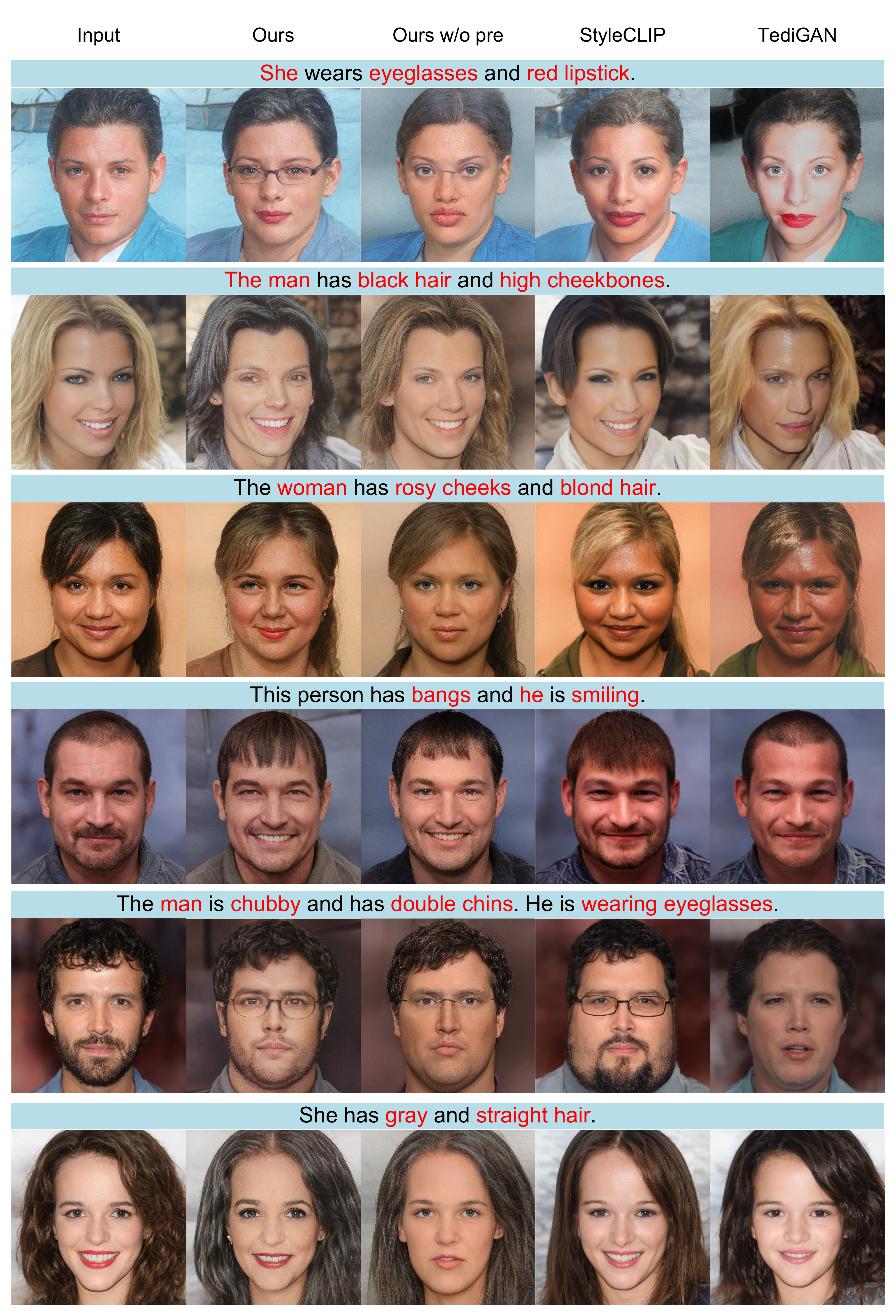}
    \caption{\textbf{Qualitative comparison of text-guided editing results}. We show from left to right the inputs, results of our full model, results of our model without pre-training, results of StyleCLIP \cite{patashnik2021styleclip} and TediGAN \cite{xia2021tedigan}. The comparison shows that our pre-trained model can benefit the downstream text-guided editing task and achieve state-of-the-art performance. Due to the vision-aware text embedding learned from pre-training, our full model is able to attend to ``She", ``eyeglasses" and ``rep lipstick" in the first case and accurately edit the images accordingly.}
    \label{fig:supp-edit}
\end{figure*}

\begin{figure*}
    \centering
    \includegraphics[width=0.8\linewidth]{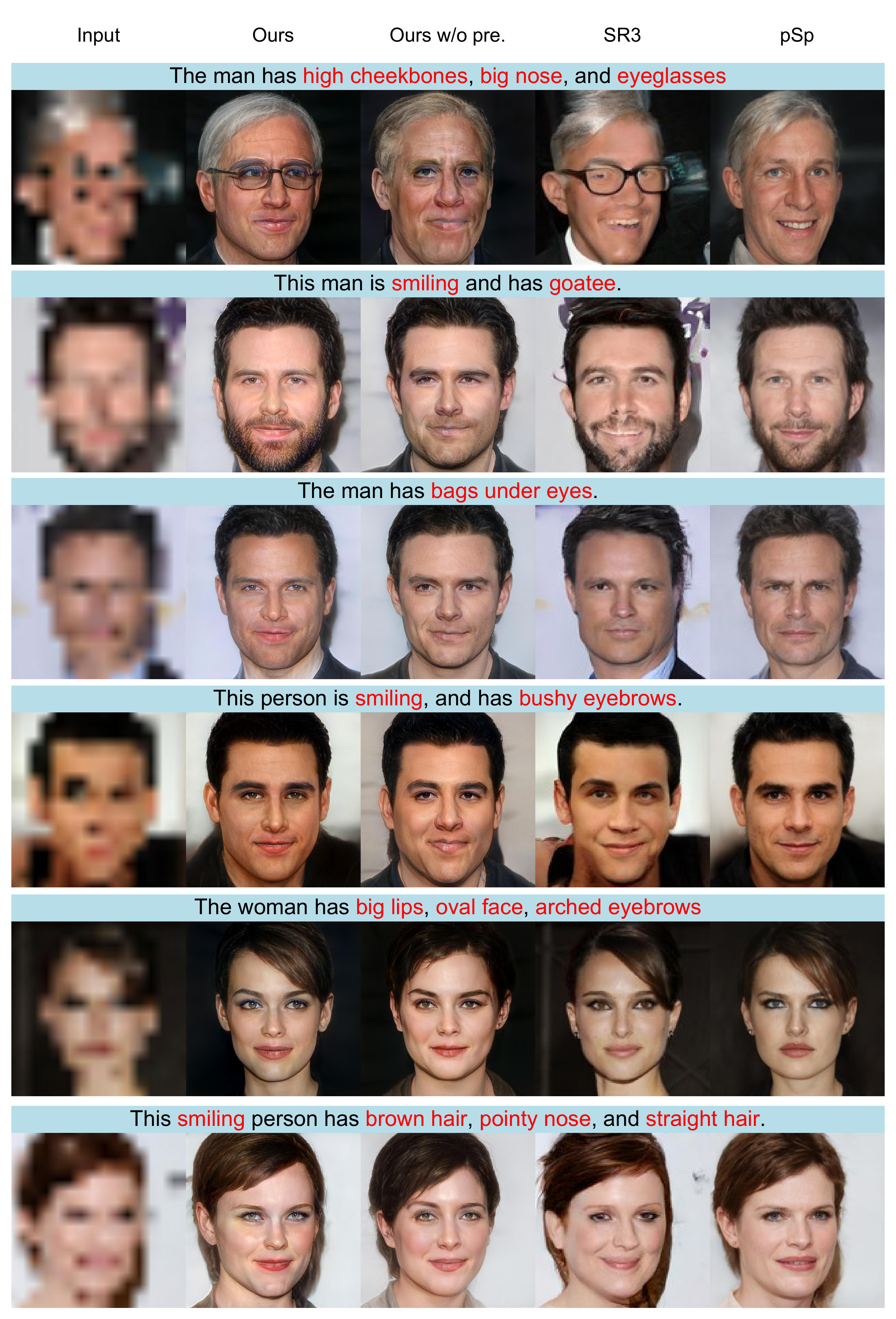}
    \caption{\textbf{Qualitative comparison of more super-resolution results}.  We show from left to right the inputs, results of our full model, results of our model without pre-training, results of  SR3~\cite{saharia2021sr3} and pSp~\cite{richardson2021psp}. Our pre-trained model could generate realistic results with more textual attributes (\textit{e.g.}, eyeglasses in the first example, big lips and arched eyebrows in the 5-th example) due to the power of our pre-trained model.}
    \label{fig:supp-sr}
\end{figure*}

\subsubsection{Experiments}
\label{sec:exp}

\paragraph{Text-Guided Editing}
To demonstrate the effectiveness of our text-guided generation framework, we show the qualitative comparison of text-guided editing results in Figure \ref{fig:supp-edit}. Specifically, we compare our full model with the one without pre-training, StyleCLIP \cite{patashnik2021styleclip} and TediGAN \cite{xia2021tedigan}. StyleCLIP and TediGAN are two state-of-the-art text-guided editing approaches. Both of them combine the strong generative powers of StyleGAN with text input for editing. Specifically, StyleCLIP maps a text prompt into an input-agnostic direction in StyleGAN's style space \cite{patashnik2021styleclip}, and TediGAN proposes to map the text into StyleGAN's style space directly. We use the released code provided by the authors on their official homepage to obtain the results in Figure \ref{fig:supp-edit}. 

The results in Figure \ref{fig:supp-edit} show that our pre-trained model can benefit the downstream text-guided editing task and achieve state-of-the-art performance. Take the first case as an example, our model without pre-training tends to make the lips bigger when it wears the lipstick, and StyleCLIP and TediGAN fail to attend to the keyword ``eyeglasses'' in the natural but relatively complex descriptions. Thanks to the power of our pre-trained model, our generation framework is able to attend to multiple attributes and edit the images accurately. 

We also provide \textbf{a video demo} generated by our full model in this supplementary material (\textbf{video.mp4}). The video demo consists of 10 video cases. In each case, we show the input on the left, our result on the right, and the input description on top. We take as input a video clip and a target description as input and generate the videos frame-by-frame. The video demo shows that our model shows promising text-guided video editing performance.

\paragraph{Text-Guided Super-Resolution}
The results of the super-resolution task are presented in Figure~\ref{fig:supp-sr}. We take relative low resolution images ($16\times 16$) as input(1-st column) and generate high-resolution results (2-nd column). We train our framework from scratch and present the results in 3-rd column, which fail to capture some textual information. Two other strong baselines we adopt are SR3~\cite{saharia2021sr3} and pSp~\cite{richardson2021psp}. Their results are presented in the 4-th and 5-th colomn respectively. Our pre-trained model could generate realistic results with more textual attributes (e.g., eyeglasses in the first example, big lips and arched eyebrows in the 5-th example).  
Super-resolution is an ill-posed problem, which means a low-resolution image may be downsampled from different high-resolution images. The details can't be well constructed with our text. How to keep the consistency between consensus frames and save the details (\textit{e.g.}, hair) are still worth exploration. Besides, the pre-trained and fixed StyleGAN~\cite{karras2019stylegan} are trained on a dataset with specific distribution and may introduce bias. With our general and diverse data, we hope we could alleviate the problem in the future.

\subsection{Ablation Study on Data Domain}

\begin{table}[h]
\footnotesize
\setlength\tabcolsep{4pt}
    \centering
    \begin{tabular}{l c r r r r} 
    \hline
    Methods  & Steps & R@1 $\uparrow$ & R@5 $\uparrow$ & R@10 $\uparrow$ & MedR $\downarrow$ \\
    \hline\hline
    HowTo100M [37] & - & 8.2  & 24.5 & 35.3  & 24.0  \\ 
    Ours (HowTo100M) & 145K & 15.7 & 38.3 & 51.3  & 10.0 \\
    Ours (HD-VILA-100M) & 145K & 6.6 & 19.5 & 27.6  & 37.0  \\
    Ours (HD-VILA-100M) & 504K & 9.1 & 25.5 & 37.3  & 20.0  \\
    \hline
    \end{tabular}
    \caption{\footnotesize Comparison of pre-training datasets on YouCook2 retrieval}
    \label{tab:rebut_2}
\end{table}

We conduct text-to-video retrieval task on YouCook2~\cite{Zhou2018TowardsAL} to check whether pre-training with in-domain dataset could benefit downstream tasks. We can see that although our model pre-trained on HD-VILA-100M outperforms HowTo100M model in their paper, our model pre-trained on HowTo100M performs best in limited epochs. This shows pre-training on in-domain dataset could benefit VL tasks very much.

\section{Datasheet for HD-VILA-100M}
In this section, we provide a DataSheet~\cite{gebru2021datasheets} for HD-VILA-100M.

\subsection{Motivation}

\begin{itemize}

\item \textbf{For what purpose was the dataset created?}
We provide this dataset in order to explore multi-modality representation learning with large scale of video-language data available in the Internet. Previous datasets are limited in scale and diversity. A large-scale video-language dataset is crucial for the research community.

\item \textbf{Who created the dataset (e.g., which team, research group) and on behalf of which entity (e.g., company, institution, organization)?}
This dataset was created by Microsoft Research Asia.

\end{itemize}

\subsection{Composition}

\begin{itemize}

\item \textbf{What do the instances that comprise the dataset represent?} 
The instances of this dataset are video and each video is paired with ASR transcripts aligned over time.

\item \textbf{How many instances are there in total?}
We include 3.3 million videos. Altogether, we extracted 103 million video clips with ASR transcripts from this data.

\item \textbf{Does the dataset contain all possible instances or is it a sample (not necessarily random) of instances from a larger set?}
It is a sample. We only keep videos with quality higher or equal to 720p from the YouTube website. The dataset covers 15 popular categories with a wide range of topics from YouTube to make it more representative.

\item \textbf{What data does each instance consist of?} 
The instance consist of a short video clip with an average duration of 13.4 seconds and an ASR transcript with 32.5 words in average.

\item \textbf{Is there a label or target associated with each instance?}
We use ASR transcripts as the labels of video clips in this dataset.

\item \textbf{Is any information missing from individual instances?}
No.

\item \textbf{Are relationships between individual instances made explicit?}
Not applicable. The relationship between videos is not the focus in our study, though it could be possible for future work.

\item \textbf{Are there recommended data splits?} 
No. We build this dataset only for pre-training so we have not created validation set this time.

\item \textbf{Are there any errors, sources of noise, or redundancies in the dataset?} 
Yes. The ASR transcripts are often noisy with mistakes. Although we use some methods to clean the data, there are still errors we cannot fix.

\item \textbf{Is the dataset self-contained, or does it link to or otherwise rely on external resources?}
The dataset is self-contained. However, we plan to only release the URLs of videos and the code for preparing data. This can protect user privacy in case some videos will be deleted by YouTube users.

\item \textbf{Does the dataset contain data that might be considered confidential?}
No. We only contain videos that are public to everyone on YouTube.

\item \textbf{Does the dataset contain data that, if viewed directly, might be offensive, insulting, threatening, or might otherwise cause anxiety?}
Yes, some videos in the YouTube are. We try our best to decrease the number of offensive videos by avoiding offensive topics.

\item \textbf{Does the dataset identify any subpopulations (e.g., by age, gender)?} 
Not explicitly (e.g., through labels).

\item \textbf{Is it possible to identify individuals, either directly or indirectly from the dataset?} 
Yes, our data includes celebrities, or other YouTube-famous people. All of the videos that we use are of publicly available data, following the Terms of Service that users agreed to when uploading to YouTube.

\item \textbf{Does the dataset contain data that might be considered sensitive in any way?}
Yes, some of YouTube videos might be. We try to avoid this by removing sensitive topics.

\end{itemize}

\subsection{Collection Process}

\begin{itemize}

\item \textbf{How was the data associated with each instance
acquired?} 
The dataset is directly observable from YouTube.

\item \textbf{What mechanisms or procedures were used to collect the data?} 
We collect the dataset using YouTube API and youtube-dl tool.

\item \textbf{If the dataset is a sample from a larger set, what was the sampling strategy?}

We use a probabilistic sampling strategy to cover more categories and make the dataset more balanced. More details can be found in \textbf{Section 3 Dataset} in the main paper.

\item \textbf{Who was involved in the data collection process and how were they compensated?}
Yuchong Sun, Bei Liu, Huan Yang, Jianlong Fu are mainly responsible for data collection. The other authors are also involved in discussing the data collection process.

\item \textbf{Over what timeframe was the data collected?}
This dataset was collected from September 2021 to October 2021, although the YouTube are often much older (dating back
to when the platform was first created).

\item \textbf{Were any ethical review processes conducted ?} 
There is no official processes conducted, since we create this dataset for research without human subjects. 

\end{itemize}

\subsection{Preprocessing/cleaning/labeling}

\begin{itemize}

\item \textbf{Was any preprocessing/cleaning/labeling of the data done?}
Yes. We process the ASR transcriptions and cut the videos into clips. More details can be found in \textbf{Section 3 Dataset} of the main paper.

\item \textbf{Was the ``raw'' data saved in addition to the preprocessed/cleaned/labeled data (e.g., to support unanticipated future uses)?}
Yes, but we do not plan to release the ``raw'' data due to copyright and privacy concerns.

\item \textbf{Is the software that was used to preprocess/clean/label the data available?}
Yes. We use an off-the-shelf tool to process ASR transcriptions, it can be found at here \footnote{\url{https://github.com/ottokart/punctuator2}}. The other code used for processing the data will also be released.

\end{itemize}

\subsection{Uses}

\begin{itemize}

\item \textbf{Has the dataset been used for any tasks already?} If so, please provide a description.
At the time of data release, only our paper has used it.

\item \textbf{Is there a repository that links to any or all papers or systems that use the dataset?}
No.

\item \textbf{What (other) tasks could the dataset be used for?}
This dataset can be used for general video-language pre-training and the pre-trained model can be transferred to a wide range of downstream tasks, e.g., video-text retrieval, video QA, video captioning.

\item \textbf{Is there anything about the composition of the dataset or the way it was collected and preprocessed/cleaned/labeled that might impact future uses?} 
Since we only release the URLs of the videos, there might be some videos missing in the future due to deleting by YouTube users or YouTube website.

\item \textbf{Are there tasks for which the dataset should not be used?}
This dataset is created for research instead of commercial usage. Tasks that are sensitive or offensive should not use this dataset.

\end{itemize}

\subsection{Distribution}

\begin{itemize}

\item \textbf{Will the dataset be distributed to third parties outside of the entity (e.g., company, institution, organization) on behalf of which the dataset was created?}
We will release the dataset to public.

\item \textbf{How will the dataset will be distributed?}
The dataset will be distributed in GitHub \footnote{\url{https://github.com/microsoft/XPretrain/tree/main/hd-vila-100m}}. We will only release the URLs of the videos and some meta-data (e.g., time span of video clips).

\item \textbf{When will the dataset be distributed?}
The dataset will be released by March 28, 2022.

\item \textbf{Will the dataset be distributed under a copyright or other intellectual property (IP) license, and/or under applicable terms of use (ToU)?} 
The dataset is under the Open Use of Data Agreement (O-UDA) \footnote{\url{https://github.com/microsoft/Open-Use-of-Data-Agreement}}.

\item \textbf{Have any third parties imposed IP-based or other restrictions on the data associated with the instances?}
No.

\item \textbf{Do any export controls or other regulatory restrictions apply to the dataset or to individual instances?} 
No.

\end{itemize}

\subsection{Maintenance}

\begin{itemize}

\item \textbf{Who will be supporting/hosting/maintaining the dataset?}
All the corresponding authors of this work.

\item \textbf{How can the owner/curator/manager of the dataset be contacted?}
By emailing the contact persons in the release page.

\item \textbf{Is there an erratum?}
No.

\item \textbf{Will the dataset be updated?}
We do not plan to update it at this time.

\item \textbf{Will older versions of the dataset continue to be supported/hosted/maintained?} 
This is the first version of this dataset.

\item \textbf{If others want to extend/augment/build on/contribute to the dataset, is there a mechanism for them to do so?} 
No at this time.

\end{itemize}

\end{document}